\documentclass{article}
\usepackage{arxiv}

\usepackage[utf8]{inputenc} 
\usepackage[T1]{fontenc}    
\usepackage{url}            
\usepackage{booktabs}       
\usepackage{amsfonts}       
\usepackage{nicefrac}       
\usepackage{microtype}      
\usepackage{lipsum}
\usepackage{algorithm,algpseudocode}
\usepackage{multirow}
\usepackage{fancyhdr}
\usepackage{multicol}
\usepackage{adjustbox}      
\usepackage{filecontents}
\usepackage{color,soul}
\usepackage{natbib}
\bibpunct[, ]{(}{)}{,}{a}{}{,}%
\usepackage[colorlinks=true,linkcolor=blue,citecolor=blue]{hyperref}

\usepackage{caption}
\usepackage{amssymb}
\usepackage{amsmath,amsthm,graphicx}
\usepackage{enumerate}
\usepackage{verbatim}
\usepackage[english]{babel}
\usepackage[table,xcdraw]{xcolor}
\usepackage{float}
\usepackage{multirow}
\usepackage[normalem]{ulem}
\usepackage{ifthen}
\usepackage{todonotes}
\usepackage{array, makecell}
\usepackage{tabu}

\newcommand{\adddavood}[1]{{\color{black} {#1}}}
\newcommand{\addafshin}[1]{{\color{black} {#1}}}

\title{A Review of Cooperative Multi-Agent Deep Reinforcement Learning}
\author{Afshin Oroojlooy and Davood Hajinezhad\\
	{\tt\small \{afshin.oroojlooy, davood.hajinezhad\}@sas.com}\\
{ SAS Institute Inc., Cary, NC, USA}
}

\usepackage{natbib}
\usepackage{graphicx}

\DeclareMathOperator*{\argmax}{arg\,max}

\newcommand{\Tau}{\mathcal{T}}

\begin{document}

\maketitle
\renewcommand{\baselinestretch}{1.3}
\large

\begin{abstract}
    Deep Reinforcement Learning has made significant progress in multi-agent systems in recent years. In this review article, we have focused on presenting recent approaches on Multi-Agent Reinforcement Learning (MARL) algorithms. In particular, we have focused on five common approaches on modeling and solving cooperative multi-agent reinforcement learning problems: (I) independent learners, (II) fully observable critic, (III) value function factorization, (IV) consensus, and (IV) learn to communicate. First, we elaborate on each of these methods, possible challenges, and how these challenges were mitigated in the relevant papers. If applicable, we further make a connection among different papers in each category. Next, we cover some new emerging research areas in MARL along with the relevant recent papers. Due to the recent success of MARL in real-world applications, we assign a section to provide a review of these applications and corresponding articles.
    Also, a list of available environments for MARL research is provided in this survey. Finally, the paper is concluded with proposals on the possible research directions.

\end{abstract}

\section{Introduction}
	Multi-Agent Reinforcement Learning (MARL) algorithms are dealing with systems consisting of several agents (robots, machines, cars, etc.) which are interacting within a common environment. Each agent makes a decision in each time-step and works along with the other agent(s) to achieve an individual predetermined goal. The goal of MARL algorithms is to learn a policy for each agent such that all agents together achieve the goal of the system. 
	Particularly, the agents are learnable units that aim to learn an optimal policy on the fly to maximize the long-term cumulative discounted reward through the interaction with the environment.
	Due to the complexities of the environments or the combinatorial nature of the problem, training the agents is typically a challenging task and several problems which MARL deals with them are categorized as NP-Hard problems, e.g. manufacturing scheduling \citep{gabel2007successful, dittrich2020cooperative}, vehicle routing problem \citep{silva2019reinforcement, zhang2020multi}, some multi-agent games \citep{bard2020hanabi} are only a few examples to mention.
	
	With the motivation of recent success on deep reinforcement learning (RL)---super-human level control on Atari games \citep{mnih2015human}, mastering the game of Go \citep{silver2016mastering}, chess \citep{silver2017mastering}, robotic \citep{kober2013reinforcement}, health care planning \citep{liu2017deep}, power grid \citep{glavic2017reinforcement}, routing \citep{nazari2018reinforcement}, and inventory optimization \citep{oroojlooyjadid2017deep}---on one hand, and the importance of multi-agent system \citep{wang2016towards, leibo2017multi} on the other hand,  several researches have been focused on deep MARL. One naive approach to solve these problems is to convert the problem to a single-agent problem and make the decision for all the agents using a centralized controller. However, in this approach, the number of actions typically exponentially increases, which makes the problem intractable. Besides, each agent needs to send its local information to the central controller and with increasing the number of agents, this approach becomes very expensive or impossible. In addition to the communication cost, this approach is vulnerable to the presence of the central unit and any incident that results in the loss of the network. Moreover, usually in multi-agent problems, each agent accesses only some local information, and due to privacy issues, they may not be allowed to share their information with the other agents.

	There are several properties of the system that is important in modeling a multi-agent system: (i) centralized or decentralized control, (ii) fully or partially observable environment, (iii) cooperative or competitive environment. %
	Within a centralized controller, a central unit takes the decision for each agent in each time step. On the other hand, in the decentralized system, each agent takes a decision for itself. 
	Also, the agents might cooperate to achieve a common goal, e.g. a group of robots who want to identify a source or they might compete with each other to maximize their own reward, e.g. the players in different teams of a game. In each of these cases, the agent might be able to access the whole information and the sensory observation (if any) of the other agents, or on the other hand, each agent might be able to observe only its local information. 
	In this paper, we have focused on the decentralized problems with the cooperative goal, and most of the relevant papers with either full or partial observability are reviewed.
	Note that \cite{marl_1995, matignon2012independent,bucsoniu2010multi, bu2008comprehensive} provide reviews on cooperative games and general MARL algorithms published till 2012. Also, \cite{da2019survey} provide a survey over the utilization of transfer learning in MARL. \cite{zhang2019multi} provide a comprehensive overview on the theoretical results, convergence, and complexity analysis of MARL algorithms on Markov/stochastic games and extensive-form games on competitive, cooperative, and mixed environments. In the cooperative setting, they have mostly focused on the theory of consensus and policy evaluation. In this paper, we did not limit ourselves to a given branch of cooperative MARL such as consensus, and we tried to cover most of the recent works on the cooperative Deep MARL. 
	In \cite{nguyen2020deep}, a  review is provided for MARL, where the focus is on deep MARL from the following perspectives: non-stationarity, partial observability, continuous state and action spaces, training schemes, and transfer learning.
	We provide a comprehensive overview of current research directions on the cooperative MARL under six categories, and we tried our best to unify all papers through a single notation.
	Since the problems that MARL algorithms deal with, usually include large state/action spaces and, the classical tabular RL algorithms are not efficient to solve them, we mostly focus on the approximated cooperative MARL algorithms. 

	The rest of the paper is organized as the following: in section \ref{sec:taxonomy}, we discuss the taxonomy and organization of the MARL algorithms we reviewed.
	in Section \ref{sec:single_agent_rl} we briefly explain the single agent RL problem and some of its components. Then the multi-agent formulation is represented and some of the main challenges of multi-agent environment from the RL viewpoint is described in Section \ref{sec:multi_agent_rl}. Section \ref{sec:iql} explains the independent Q-learner type algorithm, Section \ref{sec:centralized_critic} reviews the papers with a fully observable critic model, Section \ref{sec:value_decomposition} includes the value decomposition papers,  Section \ref{sec:consensus} explains the consensus approach, Section \ref{sec:learn_to_communicate} reviews the learn-to-communicate approach, Section \ref{sec:emerging_topics} explains some of the emerging research directions, Section \ref{sec:application} provides some applications of the multi-agent problems and MARL algorithms in real-world, Section \ref{sec:environments} very briefly mentions some of the available multi-agent environments, and finally Section \ref{sec:conclusion} concludes the paper.
	\section{Taxonomy}\label{sec:taxonomy}
	\adddavood{
In this section, we provide a high-level explanation about the taxonomy and the angle we looked at the MARL.

   A simple approach to extend single-agent RL algorithms to multi-agent algorithms is to consider each agent as an independent learner. In this setting, the other agents' actions would be treated as part of the environment. This idea was formalized in \cite{tan1993multi} for the first time where the Q-learning algorithm was extended for this problem which is called \emph{independent Q-Learning (IQL)}. The biggest challenge for IQL is {\it non-stationarity}, as the other agents' actions toward local interests will impact the environment transitions. 
		
	To address the non-stationarity issue, one strategy is to assume that all critics observe the same state (global state) and actions of all agents, which we call it \emph{fully observable critic} model. In this setting, the critic model learns the true state-value and when is paired with the actor can be used toward finding the optimal policy. When the reward is shared among all agents, only one critic model is required; however, in the case of private local reward, each agent needs to train a local critic model for itself.
		 
    Consider multi-agent problem settings, where the agents aim to maximize a single joint reward or assume it is possible to reduce the multi-agent problem into a single agent problem. 
    A usual RL algorithm may fail to find the global optimal solution in this simplified setting. 
	The main reason is that in such settings, the agents do not know the true share of the reward for their actions and as a result, some agents may get lazy over time \citep{sunehag2018value}.
	In addition, exploration among the agents with poor policy aggravates the team reward, thus restrain the agents with good policies to proceed toward optimal policy. One idea to remedy this issue is to figure out the share of each individual agent into the global reward. This solution is formalized as the {\it Value Function Factorization}, where a decomposition function is learned from the global reward. 
		 
	Another drawback in the fully observable critic paradigm is the communication cost. In particular, with increasing the number of learning agents, it might be a prohibitive task to collect all state/action information in a critic due to communication bandwidth and memory limitations. The same issue occurs for the actor when the observation and actions are being shared. Therefore, the question is how to address this communication issue and change the topology such that the local agents can cooperate and communicate in learning optimal policy. The key idea is to put the learning agents on a sparsely connected network, where each agent can communicate with a small subset of agents. Then the agents seek an optimal solution under the constraint that this solution is in \emph{consensus} with its neighbors. Through communications, eventually, the whole network reaches a unanimous policy which results in the optimal policy. 
    
    For the consensus algorithms or the fully observable critic model, it is assumed that the agent can send their observation, action, or rewards to the other agents and the hope is that they can learn the optimal policy by having that information from other agents. But, one does not know the true information that is required for the agent to learn the optimal policy. In other words, the agent might be able to learn the optimal policy by sending and receiving a simple message instead of sending and receiving the whole observation, action, and reward information. So, another line of research which is called, \emph{Learn to Communicate}, allows the agents to learn what to send, when to send, and send that to which agents. In particular, besides the action to the environment, the agents learn another action called communication action. 
     
     Despite the fact that the above taxonomy covers a big portion of MARL, there are still some algorithms that either do not fit in any of these categories or are at the intersection of a few of them. In this review, we discuss some of these algorithms too. 
     
     In Table \ref{tab:tax}, a summary of different categories are presented. Notice that in the third column we provide only a few representative references. More papers will be discussed in the  following sections. 
    
     }

\begin{table}[htbp]
    \begin{tabu} to 16.5cm {X[1.95cm,m,l]|X[3cm,p,c]|X[2cm,p,l]}
       \hline
       \textbf{Categories} & \textbf{Description} & \textbf{References}\\\hline
       Independent Learners & Each learning agent is an independent learner without considering its influence into the environment  of other agents &  \cite{tan1993multi}, \cite{fuji2018deep}, \cite{tampuu2017multiagent}, \cite{foerster2017stabilising}
\\
        \hline
       Fully Observable Critic  & All critic  models observe the same state (global state) in order to address the non-stationarity issue   & \cite{lowe2017multi}, \cite{ryu2018multi}, \cite{mao2018modelling} \\
        \hline
      Value Function Factorization & Distinguish the share of each individual agent into the global reward to avoid having lazy agents & \cite{sunehag2018value}, \cite{rashid2018qmix}, \cite{son2019qtran}\\
        \hline
       Consensus   & To avoid communication overhead, agent communicate through sparse communication network and try to reach consensus &\cite{macua2017diff}, \cite{zhang2018fully}, \cite{cassano2018multi} \\
        \hline
        Learn to Communicate    &  To improve the communication efficiency,  we allow the agents to learn what to send, when to send, and send that to which agents & \cite{foerster2016learning}, \cite{jorge2016learning}, \cite{mordatch2018emergence} \\
        \hline
   \end{tabu}
   \caption{ A summary of the MARL taxonomy in this paper.  }
   \label{tab:tax}
   \end{table}

	\section{Background, Single-Agent RL Formulation, and Multi-Agent RL Notation}\label{sec:background}
	In this section, we first go over some background of reinforcement learning and the common approaches to solve that for the single-agent problem in Section~\ref{sec:single_agent_rl}. Then, in Section~\ref{sec:multi_agent_rl} we introduce the notation and definition of the Multi-agent sequential decision-making problem and the challenges that MARL algorithms need to address.
	
	\subsection{Single Agent RL}\label{sec:single_agent_rl}
	RL considers a sequential decision making problem in which an agent interacts with an environment. The agent at time period $t$ observes state $s_t \in \mathcal{S}$ in which $\mathcal{S}$ is the state space, takes action $a_t \in \mathcal{A}(s_t)$ where $\mathcal{A}(s_t)$ is the valid action space for state $s_t$, and executes that in the environment to receive reward $r(s_t,a_t,s_{t+1}) \in \mathbb{R}$ and then transfer to the new state $s_{t+1} \in \mathcal{S}$. The process runs for the stochastic $T$ time-steps, where an episode ends. Markov Decision Process (MDP) provides a framework to characterize and study this problem where the agent has full observability of the state. 
	
	The goal of the agent in an MDP is to determine a policy $\pi: \mathcal{S} \rightarrow \mathcal{A}$, a mapping of the state space $\mathcal{S}$ to the action space $\mathcal{A}$ \footnote{A policy can be deterministic or stochastic. Action $a_t$ is the direct outcome of a deterministic policy, i.e., $a_t=\pi(s_t)$. In stochastic policies, the outcome of the policy is the probability of choosing each of the actions, i.e., $\pi(a|s_t) = Pr(a_t = a | s_t)$, and then an additional method is required to choose an action among them. For example, a greedy method chooses the action with the highest probability. In this paper, when we refer to an action resulted from a policy we mostly use the notation for the stochastic policy, $a \sim \pi(.|s)$.}, that maximizes the long-term cumulative discounted rewards:
    \begin{align}
       J = \mathbb{E}_{\pi, s_0} \left[\sum_{t=0}^{\infty}\gamma^{t}r(s_t,a_t,s_{t+1})|a_t =\pi(.|s_t)\right],
    \end{align}
	where, $\gamma\in[0,1]$ is the discounting factor. Accordingly, the value function starting from state $s$ and following policy $\pi$ denoted by $V^\pi(s)$ is given by
    \begin{align}
       V^\pi(s)=\mathbb{E}_\pi\left[\sum_{t=0}^{\infty}\gamma^{t}r(s_{t},a_{t},s_{t+1}) | a_t \sim \pi(.|s_t), s_0 = s\right],
    \end{align}
    and given action $a$, the Q-value is defined as
    \begin{align}
       Q^\pi(s,a)=\mathbb{E}_\pi\left[\sum_{t=0}^{\infty}\gamma^{t}r(s_{t},a_{t},s_{t+1}) | a_t \sim \pi(.|s_t), s_0 = s, a_0 = a\right],
    \end{align}
	Given a known state transition probability distribution $p(s_{t+1}|s_t, a_t)$ and reward matrix $r(s_t, a_t)$, Bellman \citep{bellman1957markovian} showed that the following equation holds for all state $s_t$ at any time step $t$, including the optimal values too:
	\begin{equation}
	\label{eq:bellman_expectation}
	    V^{\pi}(s_t) =  \sum_{a\in\mathcal{A}(s_t)} \pi(a|s_t) \sum_{s'\in\mathcal{S}}  p(s'|s_t, a)\left[ r(s_t, a) + \gamma V^{\pi}(s') \right],
	\end{equation}
	where $s'$ denotes the successor state of $s_t$; which will be used interchangeably with $s_{t+1}$ throughout the paper.
	Thorough maximizing over the actions, the optimal state-value and optimal policy can be obtained:
	\begin{equation}
	\label{eq:bellman_maximum}
	    V^{\pi^{\ast}}(s_t) =  \max_{a} \sum_{s'}  p(s'|s_t, a)\left[ r(s_t, a) + \gamma V^{\pi^{\ast}}(s') \right].
	\end{equation}
	Similarly, the optimal Q-value for each state-action can be obtained by: 
	\begin{equation}
	\label{eq:bellman_q_value}
	    Q^{\pi^{\ast}}(s_t, a_t) =  \sum_{s'}  p(s'|s_t, a_t)\left[ r(s_t, a_t) + \gamma \max_{a'} Q^{\pi^{\ast}}(s',a') \right].
	\end{equation}
	One can obtain an optimal policy $\pi^*$ through learning directly $Q^{\pi^{\ast}}(s_t, a_t)$. The relevant methods are called {\it value-based} methods. 
	However, in the real world usually the knowledge of the environment i.e., $p(s'|s_t, a_t)$ is not available and one cannot obtain the optimal policy using \eqref{eq:bellman_maximum} or \eqref{eq:bellman_q_value}. In order to address this issue, learning the state-value, or the Q-value,  through sampling has been a common practice. This approximation requires only samples of state, action, and reward that are obtained from the interaction with the environment. In the earlier approaches, the value for each state/state-action was stored in a table and was updated through an iterative approach. The \emph{value iteration} and \emph{policy iteration} are two famous algorithms in this category that can attain the optimal policy.
	Although, these approaches are not practical for tasks with enormous state/action spaces due to the curse of dimensionality. This issue can be mitigated through \emph{function approximation}, in which parameters of a function need to be learned by utilizing supervised learning approaches. The function approximator with parameters $\theta$ results in policy $\pi_{\theta}(a|s)$. The function approximator with parameters $\theta$ can be a simple linear regression model or a deep neural network. Given the function approximator, the goal of an RL algorithm can be re-written to maximize the utility function,
	\begin{equation}
	    J(\theta) = \mathbb{E}_{a \sim \pi_{\theta}(.|s), s \sim \rho_{\pi_{\theta}}} \sum_{t=0}^{\infty} \gamma^t r(s_t,a_t;\theta),
	\end{equation}
    where the expectation is taken over the actions and the distribution for state occupancy.

In a different class of approaches, called \emph{policy-based}, the policy is directly learned which determines the probability of choosing an action for a given state. In either of these approaches, the goal is to find parameters $\theta$ to maximize utility function $J(\theta)$ through learning with sampling. We explain a brief overview of the value-based and policy-based approaches in Sections \ref{sec:value_approximation} and \ref{sec:policy_approximation}, respectively. Note that there is a large body of literature on the single-agent RL algorithms, and we only reviewed those algorithms which are actively being used in multi-agent literature too. So, to keep the coherency, we prefer not to explore advanced algorithms like soft actor-critic (SAC) \citep{haarnoja2018soft} and TD3 \citep{fujimoto2018addressing}, which are not so common in MARL literature. For more details about other single-agent RL algorithms see \cite{li2017deep} and \cite{sutton2018reinforcement}. 
	
    For all the above notations and descriptions, we assume full observability of the environment. However, in cases that the agent accesses only some part of the state, it can be categorized as a decision-making problem with partial observability. In such circumstances, MDP can no longer be used to model the problem; instead, partially observable MDPs (POMDP) is introduced as the modeling framework. This situation happens in a lot of multi-agent systems and will be discussed throughout the paper.
	
	\subsubsection{Value Approximation}\label{sec:value_approximation}

    In value approximation, the goal is to learn a function to estimate $V(s)$ or $Q(s,a)$. It is showed that with a large enough number of observation\st{s}, a linear function approximation converges to a local optimal \citep{bertsekas96,sutton2000policy}. Nevertheless, the linear function approximators are not powerful enough to capture all complexities of a complex environment. To address this issue, there has been a tremendous amount of research on non-linear function approximators and especially neural networks in recent years \citep{mnih2013playing, bhatnagar2009convergent, gao2019reduced}.
	Among the recent works on the value-based methods, the deep Q-network (DQN) algorithm \citep{mnih2015human} attracted a lot of attention due to its human-level performance on the Atari-2600 games \citep{bellemare2013arcade}, in which it only observes the video of the game. DQN utilizes the experience replay \citep{lin1992self} and a moving target network to stabilize the training. The experience replay holds the previous observation tuple $(s_t,a_t,r_t,s_{t+1}, d_t)$ in which $d_t$ determines if the episode ended with this observation. Then the approximator is trained by taking a random mini-batch from the experience replay. Utilizing the experience replay results in sample efficiency and stabilizing the training, since it breaks the temporal correlations among consecutive observations. The DQN algorithm uses a deep neural network to approximate the Q-value for each possible action $a\in\mathcal{A}$. The input of the network is a function of the state $s_t$, e.g., concatenation/average of the last four observed states. The original paper used a combination of convolutional neural network (CNN) and fully connected (FC) as the neural network approximator; although, it can be a linear or non-linear function approximator, like any combination of FC, CNN, or recurrent neural networks (RNN). The original DQN algorithm uses a function of state $s_t$ as the input to a CNN and its output is the input to an FC neural network. This neural network with weights $\theta$ is then trained by taking a mini-batch of size $m$ from the experience replay to minimize the following loss function:
	\begin{align}
	\label{eq:dqn_loss}
	    L(\theta) =& \frac1m \sum_{i=1}^m \left( y_i - Q(s_i,a_i;\theta) \right)^2, \\ 
	    y_i =& \begin{cases}
	        r_i,  & d_t=\text{True}, \\ 
	        r_i + \gamma \max_{a'} Q(s_i',a',\theta^{-}) & d_t=\text{False}, 
	        \end{cases}
	\end{align}
	where, $\theta^{-}$ is the weights of the \emph{target network} which is updated by $\theta$ every $C$ iterations. 
	Later, new DQN-based approaches were proposed for solving RL problems. For example, inspired by \cite{hasselt2010double} which proposed double Q-learning, the Double-DQN algorithm proposed in \cite{van2016deep} to alleviate the over-estimation issue of Q-values. Similarly, Dueling double DQN \citep{wang2015dueling} proposed learning a network with two heads to obtain the advantage value $A(s,a) = Q(s,a) - V(s)$ and $V(s)$ and use that to get the Q-values. 
	In addition, another extension of the DQN algorithm is proposed by combining recurrent neural networks with DQN. DRQN \citep{hausknecht2015deep} is a DQN algorithm which uses a Long Short-Term Memory (LSTM) \citep{hochreiter1997long} layer instead of a fully connected network and is applied to a partially observable environment. 
	In all these variants, usually, the $\epsilon$-Greedy algorithm is used to ensure the exploration. That is, in each time-step with a probability of $\epsilon$ the action is chosen randomly, and otherwise, it is selected greedily by taking an {\tt argmax} over the Q-values for the state. Typically, the value of $\epsilon$ is annealed during the training. \addafshin{Choosing the hyper-parameters of the $\epsilon$-greedy algorithm, target update frequency, and the experience replay can widely affect the speed and quality of the training. For example, it is shown that a large experience replay buffer can negatively affect performance. See \cite{zhang2017deeper, liu2018effects, fedus2020revisiting} for more details.}
	
	\addafshin{In the final policy, which is used  for scoring, usually the $\epsilon$ is set to zero, which results in a deterministic policy. This  mostly works well in practice; although, it might not be applicable to stochastic policies. To address this issue, softmax operator and Boltzmann softmax operator are added to DQN \citep{lipton2016efficient, pan2019reinforcement} to get the probability of choosing each action,
	\begin{align*}
	    \text{Boltzmann}\left(Q(s_t,a)\right) = \frac{e^{\beta Q(s_t, a)}}{\sum_{a \in \mathcal{A}(s_t)} e^{\beta Q(s_t, a)}}, \forall u \in  \mathcal{A}(s_t),
	\end{align*}
	in which $\beta$ is the temperature parameter to control the rate of stochasticity of actions. To use the Boltzmann softmax operator also one needs to find a reasonable temperature parameter value and as a result cannot be used to find the general optimal stochastic policy.}
	To see other extensions of the value-based algorithm and other exploration algorithms see \cite{li2017deep}.

	\subsubsection{Policy Approximation}\label{sec:policy_approximation}

	In the value-based methods, the key idea is learning the optimal value-function or the Q-function, and from there a greedy policy could be obtained. In another direction, one can parametrize the policy, make a utility function, and try to optimize this function over the policy parameter through a supervised learning process. This class of RL algorithms is called policy-based methods, which provides a probability distribution over actions.
	For example, in the policy gradient method, a stochastic policy by parameters $\theta\in\mathbb{R}^d$ is learned. First, we define 
	\begin{equation*}
	    h(s_t,a_t;\theta) = \theta^T \phi(s_t,a_t),
	\end{equation*}
	where, $\phi(s_t,a_t)\in\mathbb{R}^d$ is called the {\it feature vector} of the  state-action pair $(s_t,a_t)$. Then, the stochastic policy can be obtained by: 
	\begin{equation*}
	    \pi(a_t|s_t;\theta) = \frac{e^{h(s_t,a_t;\theta)}}{\sum_b e^{h(s_t,b;\theta)}},
	\end{equation*}
	which is the softmax function. The goal is to directly learn the parameter $\theta$ using a gradient-based algorithm. Let us define $J(\theta)$ to measure the expected value  for policy $\pi_\theta$ for trajectory $\tau = s_0, a_0, r_0, \dots, s_T, a_T, r_T$:
	\begin{equation*}
        J(\theta) = E_{\tau} \sum_{t=0}^T \gamma^t r(s_t,a_t)
	\end{equation*}
	Then the {\it policy gradient theorem} provides an analytical expression for the gradient of $J(\theta)$ as the following:
	\begin{align}\label{eq:policy_thm}
	    \nabla_{\theta} J(\theta) & = \mathbb{E}_{\tau} \left[ \nabla_{\theta} \log p (\tau ; \theta) G(\tau) \right], \\ 
	    & = \mathbb{E}_{a \sim \pi_{\theta}(.|s), s \sim \rho_{\pi_{\theta}}} \left[ \nabla_{\theta} \log \pi (a|s ; \theta) Q_{\pi_\theta}(s,a) \right].
	\end{align}
	, in which $G(\tau)$ is the return of the trajectory. Then the policy-based methods update the parameter $\theta$ as below:
	\begin{align}
	    \theta^{t+1} = \theta^t + \alpha\widehat{ \nabla_{\theta} J(\theta)},
	\end{align}
	where, $\widehat{ \nabla_{\theta} J(\theta)}$ is an approximation of the true gradient, and $\alpha$ is the learning rate.  Depends on how to estimate $\nabla_{\theta} \log p (\tau ; \theta)$, $\log \pi (a|s ; \theta)$, or $G(\tau)$, there are several variants of policy gradient algorithms. In the following, we explore some of these variants which are mostly used for MARL algorithms.
	
	REINFORCE algorithm is one of the first policy-gradient algorithms \citep{sutton2000policy}. Particularly, REINFORCE applies Monte Carlo method and uses the actual return $G_t:=\sum_{t'=t}^T\gamma^{t'} r(s_{t'},a_{t'})$  as an approximation for $G(\tau)$ in equation \eqref{eq:policy_thm}, \addafshin{and re-writes the gradient $\nabla_{\theta} \log p (\tau ; \theta)$ as $\sum_{t=0}^T \nabla_{\theta} \log \pi_{\theta}(a_t|s_t)$. This provides an unbiased estimation of the gradient; although, it has a high variance which makes the training quite hard in practice. To reduce the variance, it has been shown that subtracting a baseline $b$ from $G(\tau)$ is very helpful and is being widely used in practice. If the baseline is a function of state $s_t$, the gradient estimator will be still an unbiased estimator, which makes $V(s)$ an appealing candidate for the baseline function. Intuitively, with a positive $G_t - V(s_t)$, we want to move toward the gradients since it results in a cumulative reward which is higher than the average cumulative reward for that state, and vice versa.  There are several other extensions of REINFORCE algorithms, each tries to minimize the gradient estimation. Among them,  REINFORCE with reward-to-go and baseline usually outperforms the REINFORCE with baseline. For more details about other extensions see \cite{sutton2018reinforcement}.} 
	
	\addafshin{REINFORCE uses the actual return from a random trajectory, which might introduce a high variance into the training. In addition, one needs to wait until the end of the episode to obtain the actual cumulative discounted reward. Actor-critic (AC) algorithm extends REINFORCE by eliminating this constraint and minimizes the variance of gradient estimation. In AC, instead of waiting until the end of the episode,  a critic model is used to approximate the value of state $s_t$ by $Q(s_t,a_t) = r(s_t,a_t) + \gamma V(s_{t+1})$. As a natural choice for the baseline, picking $V(s_t)$ results in utilizing the advantage function $A(s_t,a_t) = r(s_t,a_t) + \gamma V(s_{t+1}) - V(s_t)$. The critic model is trained by calculating the TD-error $\delta_t = r(s_t,a_t) + \gamma V_w(s_{t+1}) - V_w(s_t)$, and updating $w$ by $w = w + \alpha_w \delta_t \nabla_w V_w(s_t)$, in which $\alpha_w$ is the critic's learning rate. Therefore, there is no need to wait until the end of the episode and after each time-step one can run a train-step. With AC, it is also straight-forward to train an agent for non-episodic environments.}
	
	Following this technique, several AC-based methods were proposed. Asynchronous advantage actor-critic (A3C) contains a master node connected to a few worker nodes \citep{mnih2016asynchronous}. This algorithm runs several instances of the actor-critic model and asynchronously gathers the gradients to update the weights of a master node. Afterward, the master node broadcasts the new weights to the worker node, and in this way, all nodes are updated asynchronously. Synchronous advantage actor-critic (A2C) algorithm uses the same framework but synchronously updates the weights.

	\addafshin{
	Neither REINFORCE, AC, A2C, and A3C guarantee a steady improvement over the objective function. Trust region policy gradient (TRPO) algorithm \citep{schulman2015trust} is proposed to address this issue. TRPO tries to obtain new parameters $\theta'$ with the goal of maximizing the difference between $J(\theta') - J(\theta)$, where $\theta$ is the parameter of the current policy. Under the trajectory generated from the new policy $\pi_{\theta'}$, it can be shown that
	\begin{subequations}
	\label{eq:trpo_policy_step}
	    \begin{alignat}{3}
        J(\theta') - J(\theta) &= \mathbb{E}_{\tau \sim p_{\theta'}(\tau)} \left[ \sum_{t=0}^{\infty} \gamma^t A_{\pi_{\theta}}(s_t,a_t) \right] 
	    \label{eq:trpo_policy_step_1} \\ 
	    &= \sum_{t=0}^{\infty} \mathbb{E}_{s_t \sim p_{\theta'}(s_t)} \left[ \mathbb{E}_{a_t \sim p_{\theta'}(a_t)} \left[ \frac{p_{\theta'}(a_t|s_t)}{p_{\theta}(a_t|s_t)} \gamma^t A_{\pi_{\theta}}(s_t,a_t) \right] \right], 
	    \label{eq:trpo_policy_step_2} \\
	    & \sim \sum_{t=0}^{\infty} \mathbb{E}_{s_t \sim p_{\theta}(s_t)} \left[ \mathbb{E}_{a_t \sim p_{\theta'}(a_t)} \left[ \frac{p_{\theta'}(a_t|s_t)}{p_{\theta}(a_t|s_t)} \gamma^t A_{\pi_{\theta}}(s_t,a_t) \right] \right],
	    \label{eq:trpo_policy_step_3}
	   \end{alignat}
	\end{subequations}
	where the expectation of $s_t$ in \eqref{eq:trpo_policy_step_2} is over the $p_{\theta'}(s_t)$, though we do not have $\theta'$. 
	To address this issue, TRPO approximates \eqref{eq:trpo_policy_step_2} by substituting $s_t \sim p_{\theta'}(s_t)$ with $s_t \sim p_{\theta}(s_t)$ assuming that $\pi_{\theta'}$ is close to $\pi_{\theta}$, resulting in \eqref{eq:trpo_policy_step_3}. Under the assumption $|\pi_{\theta'}(a_t|s_t) - \pi_{\theta}(a_t|s_t)| \le \epsilon, ~ \forall s_t$, \cite{schulman2015trust} show that 
	\begin{align}
	\label{eq:trpo_policy_step_4}
        J(\theta') - J(\theta)  
	    &\ge \sum_{t=0}^{\infty} \mathbb{E}_{s_t \sim p_{\theta}(s_t)} \left[ \mathbb{E}_{a_t \sim p_{\theta'}(a_t)} \left[ \frac{p_{\theta'}(a_t|s_t)}{p_{\theta}(a_t|s_t)} \gamma^t A_{\pi_{\theta}}(s_t,a_t) \right] \right] - \sum_t{2t\epsilon C}
	\end{align}	
	which $C$ is a function of $r_{\max}$ and $T$. Thus, if $\epsilon$ is small enough, TRPO guarantees monotonic improvement under the assumption of the closeness of the policies. 
	TRPO uses Kullback–Leibler divergence $D_{KL}(p_1(x)||p_2(x)) = \mathbb{E}_{x\sim p_1(x)}\left[ \log \frac{p_1(x)}{p_2(x)}\right]$ to measure the amount of changes in the policy update. Therefore, it sets  $|\pi_{\theta'}(a_t|s_t) - \pi_{\theta}(a_t|s_t)| \le \sqrt{0.5 D_{KL}(\pi_{\theta'}(a_t|s_t) || \pi_{\theta}(a_t|s_t))} \le \epsilon$ and solves a constrained optimization problem. For solving this optimization problem, TRPO approximates the objective function by using the first order term of the corresponding Taylor series. Similarly, the constraint is approximated using the second order term of the corresponding Taylor series. This results in a policy gradient update which involves calculating the inverse of the Fisher matrix ($F^{-1}$) multiplier and a specific learning to make sure that the bound $\epsilon$ is hold. Neural networks may have millions of parameters, which make it impossible to directly achieve $F^{-1}$. To address this issue, the conjugate gradient algorithm \citep{hestenes1952methods} is used. 
	
	In general, despite TRPO's benefits, it is relatively a complicated algorithm, it is expensive to run, and it is not compatible with architectures including dropout kind of noises, or parameter sharing among different networks \citep{schulman2017proximal}. To address these issues, \cite{schulman2017proximal} proposed proximal policy optimization (PPO) algorithm in which it avoids using the Fisher matrix and its computation burden. PPO defines $r_t(\theta) = \frac{\pi_{\theta'}(a_t|s_t)}{\pi_{\theta}(a_t|s_t)}$ and it obtains the gradient update by 
	\begin{equation}
	    \mathbb{E}_{\tau \sim p_{\theta}} \left[ \sum_{t=0}^{\infty} \min (r_t(\theta) A_{\pi_{\theta}}(s_t,a_t), \text{clip}(r_t(\theta), 1-\epsilon, 1+\epsilon)A_{\pi_{\theta}}(s_t,a_t)) \right],
	\end{equation}
	which in practice works as good as TRPO in most cases.
	}

	\addafshin{Despite the issue of data efficiency, policy-based algorithms provide better convergence guarantees over the value-based algorithms \citep{yang2018finite, zhang2020global, agarwal2020optimality}. This is still true with the policy gradient which utilizes neural networks as function approximation \citep{liu2019neural, wang2019neural}. In addition, compared to the value-based algorithms, policy-based approaches can be easily applied to the continuous control problem. Furthermore, for most problems, we do not know the true form of the optimal policy, i.e., deterministic or stochastic. The policy gradient has the ability to learn either a stochastic or deterministic policy; however, in value-based algorithms, one needs to know the form of the policy at the algorithm's design time, which might be unknown. This results in two benefits of the policy-gradient method over value-based methods \citep{sutton2018reinforcement}: 
	(i) when the optimal policy is a stochastic policy (like Tic-tac-toe game), policy gradient by nature is able to learn that. However, the value-based algorithms have no way of learning the optimal stochastic policy. 
	(ii) if the optimal policy is deterministic, by following the policy gradient algorithms, there is a chance of converging to a deterministic policy. 
	However, with a value-based algorithm, one does not know the true form of the optimal policy so that he cannot choose the optimal exploration parameter (like $\epsilon$ in $\epsilon$ greedy method) to be used in the scoring time.
	In the first benefit, note that one may use a softmax operator over the Q-value to provide the probabilities for choosing each action; but, the value-based algorithms cannot learn the probabilities by themselves as a stochastic policy. Similarly, one may choose a non-zero $\epsilon$ for the score time, but one does not know the optimal value for such $\epsilon$, so this method may not result in the optimal policy.
	Also, on the second benefit note that this issue is not limited to the $\epsilon$-greedy algorithm. With added soft-max operator to a value-based algorithm, we get the probability of choosing each action. Even in this setting, the algorithm is designed to get the true values for each action, and there is no known mapping of true values to the optimal probabilities for choosing each action, which does not necessarily result in 0 and 1 actions.
	Similarly, the other variants of the soft-max operator like Boltzmann softmax which uses a temperature parameter, do not help either. Although, the temperature parameter can help to get determinism; still in practice we do not if the optimal solution is deterministic to do that.} 
	
	\subsection{Multi-Agent RL Notations and Formulation}\label{sec:multi_agent_rl}
	
    We denote a multi-agent setting with tuple  $<N, \mathcal{S}, \mathcal{A}, R, P, \mathcal{O}, \gamma >$, in which $N$ is the number of agents, $\mathcal{S}$ is state space, $\mathcal{A} = \{\mathcal{A}_1, \dots, \mathcal{A}_N \}$ is the set of actions for all agents, $P$ is the transition probability among the states, $R$ is the reward function, and $\mathcal{O} = \{\mathcal{O}_1, \dots, \mathcal{O}_N \}$ is the set of observations for all agents. Within any type of the environment, we use $\boldsymbol{a}$ to denote the vector of actions for all agents, $a_{-i}$ is the set of all agents except agent $i$, $\tau_i$ represents observation-action history of agent $i$, and $\boldsymbol{\tau}$ is the observation-action of all agents. 
	Also, $\Tau$, $\mathcal{S}$, and $\mathcal{A}$ are the observation-action space, state space, and action space, respectively. 
	Then, in a cooperative problem with $N$ agents with full observability of the environment, each agent $i$ at time-step $t$ observes the global state $s_t$ and uses the local stochastic policy $\pi_i$ to take action $a_i^t$ and then receives reward $r_i^t$. If the environment is \emph{fully cooperative}, at each time step all agents observe a joint reward value $r_t$, i.e., $r_1^t= \dots= r_N^t=r^t$. If the agents are not able to fully observe the state of the system, each agent only accesses its own local observation $o_i^t$. 
	
	Similar to the single-agent case, each agent is able to learn the optimal Q-value or the optimal stochastic policy. However, since the policy of each agent changes as the training progresses, the environment becomes non-stationary from the perspective of any individual agent. Basically, $P(s' |s,a_i,\pi_1, \dots, \pi_N) \neq P(s' |s,a_i,\pi_1', \dots, \pi_N')$ when any $\pi_i \neq \pi_i'$ so that we lose the underlying assumption of MDP.
	This means that the experience of each agent involves different co-player policies, so we cannot fix them and train an agent such that any attempt to train such models results in fluctuations of the training. This makes the model training quite challenging. Therefore, the adopted Bellman equation for MARL \citep{foerster2017stabilising} (assuming the full observability) also does not hold for the multi-agent system: 
	\begin{equation}
	    \label{eq:marl_bellman}
	    Q^{\ast}_i(s,a_i | \boldsymbol{\pi}_{-i} ) = \sum_{\boldsymbol{a}_{-i}} \boldsymbol{\pi}_{-i} (\boldsymbol{a}_{-i},s) \left[ r(s,a_i,\boldsymbol{a}_{-i}) + \gamma \sum_{s'} P(s' | s,a_i,\boldsymbol{a}_{-i}) \max_{a^{'}_i} Q^{\ast}_i(s,a^{'}_i)  \right],
	\end{equation}
	 where $\boldsymbol{\pi}_{-i} = \Pi_{j \neq i} \pi_j (a_j | s)$. Due to the fact that $\boldsymbol{\pi}_{-i}$  changes over time as the policy of other agents changes, in MARL one cannot obtain the optimal Q-value using the classic Bellman equation.
	
	On the other hand, the policy of each agent changes during the training, which results in a mix of observations from different policies in the experience replay. Thus, one cannot use the experience replay without dealing with the non-stationarity. Without experience replay, the DQN algorithm \citep{mnih2015human} and its extensions can be hard to train due to the sample inefficiency and correlation among the samples. The same issue exists within AC-based algorithms which use a DQN-like algorithm for the critic. 
	Besides, in most problems in MARL, agents are not able to observe the full state of the system, which are categorized as decentralized POMDP (Dec-POMDP). Due to the partial observability and the non-stationarity of the local observations, Dec-POMDPs are even harder problems to solve and it can be shown that they are in the class of NEXP-complete problems \citep{bernstein2002complexity}. A similar equation to \eqref{eq:marl_bellman} can be obtained for the partially observable environment too. 
	
	In the Multi-agent RL, the noise and variance of the rewards increase which results in the instability of the training. The reason is that the reward of one agent depends on the actions of other agents, and the conditioned reward on the action of a single agent can exhibit much more noise and variability than a single agent's reward. Therefore, training a policy gradient algorithm also would not be effective in general.
	
	Finally, we define the following notation which is used in a couple of papers in Nash equilibrium. A joint policy $\pi^{\ast}$ defines a Nash equilibrium if and only if: 
	\begin{equation}
		\label{eq:nash_definition}
		\forall \pi_i \in \Pi_i, \forall s \in \mathcal{S}, v_i^{(\pi^{\ast}_i, \pi^{\ast}_{-i})}(s) \ge v_i^{(\pi_i, \pi^{\ast}_{-i})}(s) ; \forall i \in \{1,\dots,N\}
	\end{equation}
	in which $v_i^{(\pi_i, \pi_{-i})}(s)$ is the expected cumulative long-term return of agent $i$ in state $s$ and $\Pi_i$ is the set of all possible policies for agent $i$. Particularly, it means that each agent prefers not to change its policy if it wants to attain the long-term cumulative discounted reward. Further, if the following holds for policy $\hat{\pi}$: 
	\begin{equation*}
	v_i^{(\hat{\pi})}(s) \ge v_i^{(\pi_i)}(s),\quad\forall i \in \{1,\dots,N\},\;\; \forall \pi_i \in \Pi_i,\; \forall s \in \mathcal{S},
	\end{equation*}
	policy $\hat{\pi}$ is called Pareto-optimal. We introduce notation for Nash equilibrium only as much that we needed for representing a few famous papers on the cooperative MARL. For more details on this topic see \cite{yang2020overview}.
	
	\section{Independent Learners}\label{sec:iql}
	One of the first proposed approaches to solve the multi-agent RL problem is to treat each agent independently such that it considers the rest of the agents as part of the environment. This idea is formalized in independent Q-Learning (IQL) algorithm \citep{tan1993multi}, in which each agent accesses its local observation and the overall agents try to maximize a joint reward. Each agent runs a separate Q-learning algorithm \citep{watkins1992q} (or it can be the newer extensions like DQN \citep{mnih2015human}, DRQN \citep{hausknecht2015deep}, etc.). IQL is an appealing algorithm since (i) it does not have the scalability and communication problem that the central control method encounters with increasing the number of agents, (ii) each agent only needs its local history of observations during the training and the inference time.
	Although it fits very well to the partially observable settings, it has the non-stationarity of environment issue.
	The tabular IQL usually works well in practice for small size problems \citep{matignon2012independent, zawadzki2014empirically}; however, in the case of function approximation, especially deep neural network (DNN), it may not work very well. 
	One of the main reasons for this weak performance is the need for the experience replay to stabilize the training with DNNs \citep{foerster2017stabilising}. 
	In an extension of IQL, Distributed Q-learning \citep{lauer2000algorithm} considers a decentralized fully cooperative multi-agent problem such that all agents observe the full state of the system and do not know the actions of the other agents, although in the training time it assumes the joint action is available for all agents. The joint action is executed in the environment and it returns the joint reward that each agent receives. This algorithm updates the Q-values only when there is a guaranteed improvement, assuming that the low returns are the result of a bad exploration of the teammates. In other words, it maximizes over the possible actions for agent $i$, assuming other agents selected the local optimal action, i.e., for a given joint action $\boldsymbol{a}_t = (a^t_1, \dots, a^t_N)$, it updates the Q-values of agent $i$ by:
	\begin{equation}
	q^{t+1}_i(s,a) = 
	\begin{cases}
		q^t_i(s,a) & \text{if}\; s \neq s_t \text{ or } a \neq a_t, \\
		\max \left\{ q^t_i(s,a), r(s_t,\boldsymbol{a}_t) + \gamma \max_{a' \in A} q^t_i \left (\delta(s_t,\boldsymbol{a}_t), a' \right) \right\} & \text{otherwise},\\
	\end{cases}
	\end{equation}
    in which $q^t_i(s,a) = \max_{\boldsymbol{a}= \{a^1, \dots, a^N\}, a^i=a} Q(s,\boldsymbol{a})$ and $\delta(s_t,\boldsymbol{a}_t)$ is the environment function which results in $s_{t+1}$. Therefore, Distributed Q-learning completely ignores the low rewards that causes an overestimated Q-values. This issue besides the curse of dimensionality results in poor performance in the problems with high dimension.
    
    Hysteretic Q-learning \citep{matignon2007hysteretic} considers the same problem and tries to obtain a good policy assuming that the low return might be the result of stochasticity in the environment so that it does not ignore them as Distributed Q-Learning does. In particular, when the TD-error is positive, it updates the Q-values by the learning rate $\alpha$, and otherwise, it updates the Q-values by the learning rate $\beta < \alpha$. Thus, the model is also robust to negative learning due to the teammate explorations.
    \cite{bowling2002multiagent} also propose  to use a variable learning rate to improve the performance of tabular IQL. In another extension for the IQL, \cite{fuji2018deep} propose to train one of the agents at each time and fix the policy of other agents within a periodic manner in order to stabilize the environment. So, during the training, other agents do not change their policies and the environment from the view-point of the single agent is stationary. 

	DQN algorithm \cite{mnih2015human} utilized \emph{experience reply} and \emph{target network} and was able to attain super-human level control on most of the Atari games. The classical IQL uses the tabular version, so one naive idea could be using the DQN algorithm instead of each single Q-learner. \cite{tampuu2017multiagent} implemented this idea and was one of the first papers which took the benefit of the neural network as a general powerful approximator in an IQL-like setting. Specifically, this paper analyzes the performance of the DQN in a decentralized two-agent game for both competitive and cooperative settings. They assume that each agent observes the full state (the video of the game), takes an action by its own policy and the reward values are also known to  both agents. The paper is mainly built on the \emph{Pong} game (from Atari-2600 environment \citep{bellemare2013arcade}) in which by changing the reward function the competitive and cooperative behaviors are obtained. In the competitive version, each agent that drops the ball loses a reward point, and the opponent wins the reward point so that it is a zero-sum game. In the cooperative setting, once either of the agents drops the ball, both agents lose a reward point. The numerical results  show that in both cases the agents are able to learn how to play the game very efficiently, that is in the cooperative setting, they learn to keep the ball for long periods, and in the competitive setting, the agents learn to quickly beat the competitor.  
    
	Experience replay is one of the core elements of the DQN algorithm. It helps to stabilize the training of the neural network and improves the sample efficiency of the history of observations. However, due to the non-stationarity of the environment, using the experience replay in a multi-agent environment is problematic. Particularly, the policy that generates the data for the experience replay is different than the current policy so that the learned policy of each agent can be misleading. In order to address this issue, \cite{foerster2016learning} disable the experience replay part of the algorithm, or in \cite{leibo2017multi} the old transitions are discarded and the experience replay uses only the recent experiences. Even though these approaches help to reduce the non-stationarity of the environment, but both limit the sample efficiency. To resolve this problem, \cite{foerster2017stabilising} propose two algorithms to stabilize the experience reply in IQL-type algorithms. They consider a fully cooperative MARL with local observation-action. In the first approach, each transition is augmented with the probability of choosing the joint action. Then, during the loss calculation, the importance sampling correction is calculated using the current policy. Thus, the loss function is changed to: 
	\begin{equation}
	    L(\theta_i) = \sum_{k=1}^b \frac{\boldsymbol{\pi}^{t_{c}}_{-i} ( \boldsymbol{a}_{-i},s) }{\boldsymbol{\pi}^{t_{i}}_{-i} ( \boldsymbol{a}_{-i},s)} \left[ (y_i^{DQN} - Q(s,a_i;\theta_i))^2 \right],
	\end{equation}
	in which $\theta_i$ is the policy parameters for agent $i$, $t_c$ is the current time-step, and $t_i$ is the time of collecting $i^{\text{th}}$ sample. In this way, the effect of the transitions generated from dissimilar policies is regularized on gradients. 
	In the second algorithm, named FingerPrint, they propose augmenting the experience replay with some parts of the policies of the other agents. However, the number of parameters in DNN is usually large and as a result, it is intractable in practice. Thus, they propose to augment each instance in the experience replay by the iteration number $e$ and the $\epsilon$ of the $\epsilon$-greedy algorithm. 
	In the numerical experiments, they share the weights among the agent, while the id of each agent is also available as the input. They provide the results of two proposed algorithms plus the combination of them on the StarCraft game \citep{samvelyan19smac} and compare the results by a classic experience replay and one no-experience replay algorithm. They conclude that the second algorithm obtains better results compared to the other algorithm. 
	
	\cite{omidshafiei2017deep} propose another extension of the experience replay for the MARL. They consider multi-task cooperative games, with independent partially observable learners such that each agent only knows its own action, with a joint reward. An algorithm, called HDRQN, is proposed which is based on the DRQN algorithm \citep{hausknecht2015deep} and the Hysteretic Q-learning \citep{matignon2007hysteretic}. 
	Also, to alleviate the non-stationarity of MARL, the idea of Concurrent Experience Replay Trajectories (CERTs) is proposed, 
	 in which the experience replay gathers the experiences of all agents in any period of one episode and also during the sampling of a mini-batch, it obtains the experiences of one period of all agents together.
	Since they use LSTM, the experiences in the experience replay are zero-padded (adds zero to the end of the experiments with smaller sizes to make the size of all experiments equal). 
	Moreover, in the multi-task version of HRDQN, there are different tasks that each has its own transition probability, observation, and reward function. During the training, each agent observes the task ID, while it is not accessible in the inference time.
	To evaluate the model, a two-player game is utilized, in which agents are rewarded only when all the agents simultaneously capture the moving target. In order to make the game partially observable, a flickering screen is used such that with $30\%$ chance the screen is flickering. The actions of the agents are moving north, south, west, east, or waiting. Additionally, actions are noisy, i.e. with $10\%$ probability the agent might act differently than what it wanted.
    \section{Fully Observable Critic} \label{sec:centralized_critic}
	Non-stationarity of the environment is the main issue in multi-agent problems and MARL algorithms. One of the common approaches to address this issue is using a fully observable critic. The fully observable critic involves the observations and actions of all agents and as a result, the environment is stationary even though the policy of other agents changes. In other words, $P(s' |s,a_1, \dots, a_N,\pi_1, \dots, \pi_N) = P(s' |s,a_1, \dots, a_N,\pi_1', \dots, \pi_N')$ even if $\pi_i \neq \pi_i'$, since the environment returns an equal next-state regardless of the changes in the policy of other agents. Following this idea, there can be 1 or $N$ critic models: (i) in a fully cooperative problems, one central critic is trained, and (ii) when each agent observes a local reward, each agent may need to train its own critic model, resulting to $N$ critic models. In either case, once the critic is fully observable, the non-stationarity of critic is resolved and it can be used as a good leader for local actors. 
	
	Using this idea, \cite{lowe2017multi} propose a model-free multi-agent reinforcement learning algorithm to the problem in which agent $i$ at time step $t$ of execution accesses its own local observation $o_i^t$, local actions $a_i^t$, and local rewards $r_i^t$. They consider cooperative, competitive, and mixed competitive and cooperative games, and proposed Multi-agent DDPG (MADDPG) algorithm in which each agent trains a DDPG algorithm such that the actor $\pi_i(o_i;\theta_i)$ with policy weights $\theta_i$ observes the local observations while the critic $Q_i^{\mu}$ is allowed to access the observations, actions, and the target policies of all agents in the training time. Then, the critic of each agent concatenates all state-actions together as the input and using the local reward obtains the corresponding Q-value. Either of $N$ critics are trained by minimizing a DQN-like loss function: 
	\begin{align*}
	    L(\mu_i) = \mathbb{E}_{\boldsymbol{o}^t,a,r,\boldsymbol{o}^{t+1}} \left[ \left( Q_i(\boldsymbol{s}^t,a_1^t,\dots, a_N^t; \mu_i) - y \right)^2 \right], \\ 
	    y = r_i^t + \gamma Q_i \left(\boldsymbol{o}^t,\bar{a}^{t+1}_1,\dots, \bar{a}^{t+1}_N ; \bar{\mu}_i \right) |_{\bar{a}_j^{t+1}  = \bar{\pi}(o_j^{t+1})},
	\end{align*}
	in which $\boldsymbol{o}^t$ is observation of all agents, $\bar{\pi}_j$ is the target policy, and $\bar{\mu}$ is the target critic. As a result, the \emph{critic of each agent deals with a stationary environment}, and in the inference time, it only needs to access the local information. MADDPG is compared with the decentralized trained version of DDPG \citep{lillicrap2015continuous}, DQN \citep{mnih2015human}, REINFORCE \cite{sutton2000policy}, and TRPO \citep{schulman2015trust} algorithm in a set of  grounded communication environments from particle environment \citep{haarnoja2018soft}, e.g., predator-prey, arrival task, etc. The continues space-action predator-prey environment from this environment is usually considered as a benchmark for MARL algorithms with local observation and cooperative rewards. In the most basic version of the predator-prey, there are two predators which are randomly placed in a $5\times5$ grid, along with one prey which also randomly is located in the grid. Each predator observes its direct neighbor cells, i.e., $3\times3$ cells, and the goal is to catch the prey together to receive a reward, and in all other situations, each predator obtains a negative reward. 
    	
	Several extensions of MADDPG algorithm are proposed in the literature and we review some of them in the rest of this section. \cite{ryu2018multi} propose an actor-critic model with local actor and critic for a DEC-POMDP problem, in which each agent observes a local observation $o_i$, observes its own reward $r_i : \mathcal{S}\times\mathcal{A}_1\times\dots\times\mathcal{A}_N \rightarrow \mathbb{R}$, and learns a deterministic policy $\mu_{\theta_i} : \mathcal{O}_i \rightarrow \mathcal{A}_i$. The goal for agent $i$ is to maximize its own discounted return $R_i = \sum_{t=0}^{\infty} \gamma^t r_i^t$. An extension of MADDPG with a generative cooperative policy network, called MADDPG-GCPN, is proposed in which there is an extra actor network $\mu_i^c$ to generate action samples of other agents. Then, the critic uses the experience replay filled with sampled actions from GCPN, and not from the actor of the other agents. So, there is no need to share the target policy of other agents during the training time. Further, the algorithm is modified in a way such that the critic can use either immediate individual or joint reward during the training. 
	They presented a new version of the predator-prey game in which each agent receives an individual reward plus a shared one if they catch the prey. The experimental analysis on the predator-prey game and a controlling energy storage systems problem shows that the standard deviation of obtained Q-values is lower in MADDPG-GCPN compared to MADDPG. 
	
	As another extension of MADDPG, \cite{chu2017parameter} consider multi-agent cooperative problems with $N$ agents and proposes three actor-critic algorithms based on MADDPG. The first one assumes that all agents know the global reward and shares the weights between agents so that it \st{actually} includes one actor and one critic network. The second algorithm assumes the global reward is not shared and each agent indeed updates its own critic using the local reward so that there are $N$ critic networks. Although, agents share their weights so that there is only one actor network which means $N+1$ networks are trained. The third algorithm also assumes non-shared global reward, though uses only two networks, one actor network and one critic network such that the critic has $N$ heads in which head $i \in \{1,\dots,N\}$ provides the Q-value of the agent $i$. They compare the results of their algorithms on three new games with MADDPG, PPO \citep{schulman2017proximal}, and PS-TRPO algorithms (PS-TRPO is the TRPO algorithm \citep{schulman2015trust} with parameter sharing, see \cite{sukthankar2017autonomous} for more details).

	\cite{mao2018modelling} present another algorithm based on MADDPG for a cooperative game, called ATT-MADDPG which considers the same setting as in \cite{lowe2017multi}. It enhances the MADDPG algorithm by adding an attention layer in the critic network. In the algorithm, each agent trains a critic which accesses the actions and observations of all agents. To obtain the Q-value, an attention layer is added on the top of the critic model to determine the corresponding Q-value.   In this way, at agent $i$, instead of just using $[o_1^t, \dots, o_N^t]$ and $[a^t_i,a^t_{-i}]$ for time step $t$, ATT-MADDPG considers $K$ combinations of possible action-vector $a_{-i}^t$, and obtains the corresponding $K$ Q-values. Also, using an attention model, it obtains the weights of all $K$ action-sets such that the hidden vector $h_i^t$ of the attention model is generated via the actions of other agents ($a_{-i}^t$). Then, the attention weights are used to obtain the final Q-value by a weighted sum of the $K$ possible Q-values. Indeed, their algorithm combines the MADDPG with the k-head Q-value \citep{van2017hybrid}. They provide some numerical experiments on cooperative navigation, predator-prey, a packet-routing problem, and compare the performance with MADDPG and few other algorithms. Moreover, the effect of small or large $K$ is analyzed within each environment.   
    In \cite{wang2019r} again the problem setting is considered  as the MADDPG, though they assume a limit on the communication bandwidth. 
    Due to this limitation, the agents are not able to share all the information, e.g., they cannot share their locations on "arrival task" (from Multi-Agent Particle Environment \citep{mordatch2017emergence}), which limits the ability of the MADDPG algorithm to solve the problem. To address this issue, they propose R-MADDPG, in which a recurrent neural network is used to remember the last communication in both actor and critic. In this order, they modified the experience replay such that each tuple includes ($o_{i}^t, a_{i}^t, o^{t+1}_i, r^{t}_i, h_{i}^t, h^{t+1}_i$), in which $h_{t}^i$ is the hidden state of the actor network. The results are compared with MADDPG over the "arrival task" with a communication limit, where each agent can select either to send a message or not, and the message is simply the position of the agent. With the fully observable state, their algorithm works as well as MADDPG. On the other hand, within the partially observable environment, recurrent actor (with fully connected critic) does not provide any better results than MADDPG; though, applying both recurrent actor and recurrent critic, R-MADDPG obtains higher rewards than MADDPG. 
    
    Since MADDPG concatenates all the local observations in the critic, it faces the curse of dimensionality with increasing the number of agents. To address this issue, \cite{iqbal2018actor} proposed Multiple Actor Attention-Critic (MAAC) algorithm which efficiently scales up with the number of agents. The main idea in this work is to use an attention mechanism \cite{choi2017multi,jiang2018learning} to select relevant information for each agent during the training. In particular, agent $i$ receives the observations, $o = (o_1, ..., o_N)$, and actions, $a = (a_1, ..., a_N)$ from all agents. The value function parameterized by $\psi$, $Q^\psi_i(o, a)$,  is defined as  a function of agent $i$'s observation-action, as well as the information received from the other agents:
    \begin{align*}
    Q^\psi_i(o, a)  = f_i(g_i(o_i, a_i), x_{i}),
    \end{align*}
    where $f_i$ is a two-layer perceptron, $g_i$ is a one-layer  embedding function, and $x_i$ is the contribution of other agents. In order to fix the size of $x_i$, it is set equal to the weighted sum of other agents' observation-action:
    \begin{align*}
        x_i = \sum_{j\ne i} \alpha_j v_j =  \sum_{j\ne i} \alpha_j h( V g_j(o_j, a_j)),
    \end{align*}
    where $v_j$ is a function of the embedding of agent $j$, encoded with an embedding function and then linearly transformed by a shared matrix $V$, and $h$ is the activation function. Denoting $e_j=g_j(o_j, a_j)$, using the query-key system \citep{vaswani2017attention}, the attention weight $\alpha_j$ is proportional to:
    \begin{align*}
        \alpha_j \propto \exp ( e_j^T W_k^T W_q e_i),
    \end{align*}
    where $W_q$ transforms $e_i$ into a ``query'' and $W_k$ transforms $e_j$ into a ``key''. 
    The critic step updates the $\psi$ through minimizing the following loss function:
    \begin{align*}
    \mathcal{L}_{Q}(\psi) = \sum_{i=1}^N{\mathbb{E}_{(o, a, r, o')\sim D}\left[(Q_i^\psi(o, a) - y_i)^2\right]},    
    \end{align*}
    where,
    \begin{align*}
    y_i = r_i + \gamma \mathbb{E}_{a' \sim \pi_{\bar{\theta}}(o')}\left[Q_i^{\bar{\psi}}(o', a') - \alpha~\log(\pi_{\bar{\theta_i}}(a_i'|o_i'))\right],
    \end{align*}
    where $\bar{\psi}$ and $\bar{\theta}$ are the parameters of the target critics and target policies respectively. In order to encourage exploration, they also use the idea of soft-actor-critic  (SAC) \citep{haarnoja2018soft}. MAAC is compared with COMA \citep{foerster2018counterfactual} (will be discussed shortly), MADDPG, and their updated version with SAC, as well as an independent learner with DDPG \citep{lillicrap2015continuous}, over two environments: treasure collection and  rover-tower. MAAC obtains better results than the other algorithms such that the performance gap becomes shorter as  the number of agents increases.  
   
   \cite{jiang2018graph} assume a graph connection among the agents such that each node is an agent. A partially observable environment is assumed in which each agent observes a local observation, takes a local action, and receives a local reward, while agents can share their observations with their neighbors and the weights of all agents are shared. A graph convolutional reinforcement learning for cooperative multi-agent is proposed. The multi-agent system is modeled as a graph such that agents are the nodes, and each has some features which are the encoded local observations. A multi-head attention model is used as the convolution kernel to obtain the connection weights to the neighbor nodes. To learn Q-function an end-to-end algorithm, named DGN,  is proposed, which uses the centralized training and distributed execution (CTDE) approach. The goal is to maximize the sum of the reward of all agents. During the training, DGN allows the gradients of one agent to flow  K-neighbor agents and its receptive field to stimulate cooperation.  
	In particular, DGN consists of three phases: (i) an observation encoder, (ii) a convolutional layer, and (iii) Q-network. The encoder (which is a simple MLP, or convolution layer if it deals with images) receives observation $o_{i}^t$ of agent $i$ at time $t$ and encodes it to a feature vector $h_{i}^t$. In phase (ii), the convolutional layer integrates the local observations of $K$ neighbors to generate a latent feature ${h'}_{i}^t$. In order to obtain the latent vector, an attention model is used to make the input independent of the number of input features. The attention model gets all feature vectors of the $K$-neighbors, generates the attention weights, and then calculates the weighted sum of the feature vectors to obtain ${h'}_{i}^t$. Another convolution layer may be added to the model to increase the receptive field of each agent, such that ${h'}_{i}^t$ are the inputs of that layer, and ${h^{''}}_{i}^t$ are the outcomes. Finally, in phase (iii), the Q-network provides the Q-value of each possible action. Based on the idea of DenseNet \citep{huang2017densely}, the Q-network gathers observations and all the latent features and concatenates them for the input. (Note that this procedure is followed for each agent, and the weights of the network are shared among all agents.)
	In the loss function, besides the Q-value loss function, a penalized KL-divergence is added. This function measures the changes between the current attention weights and the next state attention weights and tries to avoid drastic changes in the attention weight. Using the trained network, in the execution time each agent $i$ observes $o_{i}^t$ plus the observation of its $K$ neighbors ($\{o_{tot}\}^t_{j \in \mathbb{N}(i)}$) to get an action. The results of their algorithm are compared with DQN, CommNet \citep{sukhbaatar2016learning}, and MeanField Q-Learning \citep{yang2018mean}, on Jungle, Battle, and Routing environments. 
	
    \cite{yang2018mean} considers the multi-agent RL problems in the case that there exists a huge number of agents collaborating or competing with each other to optimize some specific long-term cumulative discounted rewards. They propose {\it Mean Field Reinforcement Learning} framework, where every single agent only considers an average effect of its neighborhoods, instead of exhaustive communication with all other agents within the population. Two algorithms namely Mean Field Q-learning (MF-Q) and Mean Field Actor-Critic (MF-AC) are developed following the mean-field idea. There exists a single state visible to all agents, and the local reward and the local action of all agents are also visible to the others during the training. Applying Tylor' theorem, it is proved that $Q^i(s,a)$ can be approximated by $Q^i(s,a^i,\bar{a}^i)$, where $a$ concatenates the action of   all agents, $a^i$ is the action of agent $i$, and $\bar{a}^i$ denotes the average action from the neighbors. Furthermore, utilizing the contraction mapping technique, it is shown that mean-field Q-values converge to the Nash Q-values under some particular assumptions. The proposed algorithms are tested on three different problems:  Gaussian Squeeze and the Ising Model (a framework in statistical mechanics to mathematically model ferromagnetism), which are  cooperative problems, and the battle game, which is a mixed cooperative-competitive game. The numerical results show the effectiveness of the proposed method in the case of many-agent RL problems. 

	\citep{foerster2018counterfactual} proposes COMA, a model with a single centralized critic which uses the global state, the vector of all actions, and a joint reward. This critic is shared among all agents, while the actor is trained locally for each agent with the local observation-action history. 
	The joint reward is used to train $Q(s^t,[a_i^t, a_{-i}^t])$. Then, for the agent $i$, with the fixed $a^t_{-i}$ the actor uses a counterfactual baseline 
	$$b(s^t,a_{-i}^t) = \sum \limits_{\hat{a}_i^t} \pi_i(\hat{a}_i^t|o_i^t)Q(s^t,[\hat{a}_i^t, a^t_{-i}]),$$ 
	to obtain contribution of action $a_i^t$ via advantage function $A(s,a_i^t) = Q(s, [a_{-i}^t,a_i]) - b(s^t,a_{-i}^t)$. Also, each actor shares its weights with other agents, and uses gated recurrent unit (GRU) \citep{cho2014learning} to utilize the history of the observation. They present the results of their algorithm on StarCraft game \citep{samvelyan19smac} and compare COMA with central-V, central-QV, and two implementations of independent actor-critic (IAC).  
	
	\cite{yang2018cm3} consider a multi-agent cooperative problem in which in addition to the cooperative goal, each agent needs to attain some personal goals. Each agent observes a local observation, local reward corresponding to the goal, and its own history of actions, while the actions are executed jointly in the environment. This is a common situation in problems like autonomous car driving. For example, each car has to reach a given destination and all cars need to avoid the accident and cooperate in intersections. The authors propose an algorithm (centralized training, decentralized execution) called CM3, with two phases. In the first phase, one single network is trained for all agents to learn  personal goals. The output of this network, a hidden layer, is passed to a given layer of the second network to initialize it, and the goal of the second phase is to attain the global goal. Also, since the collective optimal solution of all agents is not necessarily optimal for every individual agent, a credit assignment approach is proposed to obtain the global solution of all agents. This credit assignment, motivated by \cite{foerster2018counterfactual}, is embedded in the design of the second phase.
	In the first phase of the algorithm, each agent is trained with an actor-critic algorithm as a single agent problem learning to achieve the given goal of the agent. In this order, the agent is trained with some randomly assigned goal very well, i.e., agent $i$ wants to learn the local policy $\pi$ to maximize $J_i(\pi)$ for any arbitrary given goal. All agents share the weights of the policy, so the model is reduced to maximize $J_{local}(\pi) = \sum_{i=1}^N J_i(\pi)$. Using the advantage approximation, the update is performed by $\nabla_{\theta} J_{local}(\pi) = \mathbb{E} \left[ \sum_{i=0}^{N} \nabla_{\theta} \log \pi_i(a_i | o_i, g_i) (R(s,a_i,g_i) + \gamma V(o_i^{t+1}, g_i) - V(o_i^{t}, g_i)) \right]$, in which $g_i$ is the goal of agent $i$.
	The second phase starts by the pre-trained agents, and trains a new global network in the multi-agent setting to achieve a cooperative goal by a comprehensive exploration. The cooperative goal is the sum of local rewards, i.e., $J_{global}(\pi) = \mathbb{E}_{\pi} \left[ \sum_{t=0}^{\infty} \gamma^t R_g^t \right]$, in which $R_g^t = \sum_{i=1}^{N} r(o_i^t,a_i^t, g_i) $. 
	In the first phase, each agent only observes the part of the state that involves the required observation to complete the personal task. In the second phase, additional observations are given to each agent to achieve the cooperative goal. This new information is used to train the centralized critic and also used in the advantage function to update the actor's policy. The advantage function uses the counterfactual baseline \citep{foerster2018counterfactual}, so that the global objective is updated by $\nabla_{\theta} J_{global}(\pi) = \mathbb{E} \left[ \sum_{i=0}^{N} \nabla_{\theta} \log \pi_i(a_i | o_i, g_i) (Q(s,a,g) - b(s,a^{-i},g) ) \right]$. Finally, a combined version of the local and global models is used in this phase to train the model with the centralized critic. They present the experiments on an autonomous vehicle negotiation problem and compare the results with COMA \citep{foerster2018counterfactual} and independent actor-critic learners model. 
	
	\cite{sartoretti2019distributed} extend A3C \citep{mnih2016asynchronous} algorithm for a centralized actor and a centralized critic. The proposed algorithm is based on  centralized training and decentralized execution, in a fully observable environment. They consider a construction problem, TERMES \citep{petersen2012termes}, in which each agent is responsible to gather, carry, and place some blocks to build a certain structure. Each agent observes the global state plus its own location, takes its own local action, and executes it locally (no joint action selection/execution). Each agent receives its local sparse reward, such that  the reward is +1 if it puts down a block in a correct position and -1 if it picks up a block from a correct position. Any other actions receive 0 rewards.
	During the training, all agents share the weights (as it is done in A3C) for both actor and critic models to train a central agent asynchronously. In the execution time, each agent uses one copy of the learned policy without communicating with other agents. The goal of all agents is to achieve the maximum common reward. The learned policy can be executed in an arbitrary number of agents, and each agent can see the other agents as moving elements, i.e., as part of the environment. 

    Finally, in \cite{kim2019learning} a multi-agent problem is considered under the following two assumptions: 1) The communication bandwidth among the agents is limited. 2) There exists a shared communication medium such that at each time step only a subset of agents is able to use it to broadcast their messages to the other agents. Therefore,  communication scheduling is required to determine which agents are able to broadcast their messages. Utilizing the proposed  framework, which is called SchedNet, the agents are able to schedule themselves, learn how to encode the received messages, and also learn how to pick actions based on these encoded messages. SchedNet focuses on centralized training and distributed execution. Therefore, in training the global state is available to the critic, while the actor is local for each agent and the agents are able to communicate through the limited channel. To control the communication,  a Medium Access Control (MAC) protocol is proposed, which uses a weight-based scheduler (WSA) to determine which nodes can access the shared medium. The local actor $i$ contains three networks: 1) message encoder, which takes the local observation $o_i$ and outputs the messages $m_i$. 2) weight generator, which takes the local observation $o_i$ and outputs the weight $w_i$. Specifically, the $w_i$ determines the importance of observations in node $i$. 3) action selector. This network receives the observation $o_i$, encoded messages $m_i$ plus the information from the scheduling module, which selects $K$ agents who are able to broadcast their messages. Then maps this information to the action  $a_i$. A centralized critic is used during the training to criticize the actor. In particular, the critic receives the global state of the environment and the weigh vector $W$ generated by the wight generator networks as an input, and the output will be $Q(S,W)$  as well as $V(S)$. The first one is used to update the actor wight $w_i$ while the second one is used for adjusting the weights of two other networks, namely action selector and message encoder. Experimental results on the predator-prey, cooperative-communication, and navigation task demonstrate that intelligent communication scheduling can be helpful in MARL. 
 
	\section{Value Function Factorization} \label{sec:value_decomposition}
	
	Consider a cooperative multi-agent problem in which we are allowed to share all information among the agents and there is no  communication limitation among the agents.  Further, let's assume that we are able to deal with the huge action space. In this scenario, a centralized RL approach can be used to solve the problem, i.e., all state observations are merged together and the problem is reduced to a single agent problem with a combinatorial action space. However, \cite{sunehag2018value} shows that naive centralized RL methods fail to find the global optimum, even if we are able to solve the problems with such a huge state and action space. The issue comes from the fact that some of the agents may get lazy and not learn and cooperate as they are supposed to. This may lead to the failure of  the whole system. One possible approach to address this issue is to determine the role of each agent in the joint reward and then somehow isolate its share out of it. This category of algorithms is  called {\it Value Function Factorization}.
	
	In POMDP settings, if the optimal reward-shaping is available, the problem reduces to train several independent learners, which simplifies the learning. Therefore, having a reward-shaping model would be appealing for any cooperative MARL. However, in practice it is not easy to divide the received reward among the agents since their contribution to the reward is not known or it is hard to measure. Following this idea, the rest of this section discusses the corresponding algorithms. 
    
    In the literature of tabular RL, there are two common approaches for reward shaping: (i) \emph{difference rewards} \citep{agogino2004unifying} which tries to isolate the reward of each agent from the joint reward, i.e. $\hat{r}_i = r - r_{-i}$, where $r_{-i}$ denotes the other agents' share than agent $i$ in the global reward,
    (ii) \emph{ Potential-based reward shaping} \citep{ng1999policy}. In this class of value function factorization methods, term $r + \Phi(s') - \Phi(s)$ is used instead of mere $r$, in which $\Phi(s)$ defines the desirability of the agent to be at state $s$. This approach is also extended for online POMDP settings \citep{eck2016potential} and multi-agent setting \citep{devlin2011theoretical}, though defining the potential function is challenging and usually needs specific domain knowledge. In order to address this issue, \cite{devlin2014potential} combine these approaches and proposes two tabular algorithms. 
    Following this idea, several value function factorization algorithms are proposed to automate reward-shaping and avoid the need for a field expert, which are summarized in the following.
	
	\cite{sunehag2018value} consider a fully cooperative multi-agent problem (so a single shared reward exists) in which each agent observes its own state and action history. An algorithm, called VDN, is proposed to decompose the value function for each agent. Intuitively, VDN measures the impact of each agent on the observed joint reward. 
    It is assumed that the joint action-value function $Q_{tot}$ can be additively decomposed into $N$ Q-functions for $N$ agents, in which each Q-function only relies on the local state-action history, i.e., 
    \begin{equation}
        Q_{tot} = \sum \limits_{i=1}^N Q_i(\tau_i, a_i, \theta_i) 
    \end{equation}
    In other words, for the joint observation-action history $\boldsymbol{\tau}$, it assumes validity of the individual-global-max (IGM) condition. Individual action-value functions $[Q_i: \Tau \times \mathcal{A}]_{i=1}^{N} $ satisfies IGM condition for the joint action-value function $Q_{tot}$, if: 
    \begin{equation}
        \argmax_{\boldsymbol{a}} Q_{tot}(\boldsymbol{\tau}, \boldsymbol{a}) = 
        \begin{pmatrix}
        \argmax_{a_1} Q_1 (\tau_1, a_1) \\ 
        \vdots \\ 
        \argmax_{a_N} Q_N (\tau_N, a_N) \\
        \end{pmatrix},
    \end{equation}
    in which $\boldsymbol{\tau}$ is the vector of local observation of all agents, and $u$ is the vector of actions for all agents. 
    Therefore, each agent observes its local state, obtains the Q-values for its action, selects an action, and then the sum of Q-values for the selected action of all agents provides the total Q-value of the problem. Using the shared reward and the total Q-value, the loss is calculated and then the gradients are backpropagated into the networks of all agents. 
    In the numerical experiments, a recurrent neural network with dueling architecture \citep{wang2015dueling} is used to train the model. Also, two extensions of the model are analyzed: (i) shared the policy among the agent, by adding the one-hot-code of the agent id to state input, (ii) adding information channels to share some information among the  agents. Finally, VDN is compared with independent learners, and centralized training, in  three versions of the two-player 2D grid.
	
	QMIX \citep{rashid2018qmix} considers the same problem as VDN does, and proposed an algorithm which is in fact an improvement over VDN \citep{sunehag2018value}. As mentioned, VDN adds some restrictions to have the additivity of the  Q-value and further shares the action-value function during the training. QMIX also shares the action-value function during the training (a centralized training algorithm, decentralized execution); however, adds the below constraint to the problem:
	\begin{equation}
	    \frac{\partial Q_{tot}}{\partial Q_i} \ge 0, \forall i,
	\end{equation}
	which enforces positive weights on the mixer network, and as a result, it can guarantee (approximately) monotonic improvement. Particularly, in this model, each agent has a $Q_i$ network and they are part of the general network ($Q_{tot}$) that provides the Q-value of the whole game. Each  $Q_i$ has the same structure as DRQN \citep{hausknecht2015deep}, so it is trained using the same loss function as DQN. 
		Besides the monotonicity  constraint over the relationship between $Q_{tot}$ and each $Q_i$, QMIX  adds some extra information from the global state plus a non-linearity into $Q_{tot}$ to improve the solution quality. They provide numerical results on StarCraft II and compare the solution with VDN.

	Even though VDN and QMIX cover a large domain of multi-agent problems, the assumptions for these two methods do not hold for all problems. To address this issue, \cite{son2019qtran} propose QTRAN algorithm. The general settings are the same as VDN and QMIX (i.e., general DEC-POMDP problems in which each agent has its own partial observation, action history, and all agents share a joint reward). The key idea here is that the actual $Q_{tot}$ may be different than $\sum_{i=1}^N Q_i(\tau_i, a_i, \theta_i)$. However, they consider an alternative joint action-value $Q^{'}_{tot}$, assumed to be factorizable by additive decomposition.
    Then, to fill the possible gap between $Q_{tot}$ and $Q^{'}_{tot}$ they introduce 
    \begin{equation}
        V_{tot} = \max_{\boldsymbol{a}} Q_{tot}(\boldsymbol{\tau}, \boldsymbol{a}) - \sum \limits_{i=1}^N Q_i(\tau_i, \bar{a}_i), 
    \end{equation}
     in which $\bar{a}_i$ is $\argmax_{a^{'}_{i}} Q_i(\tau_i, a^{'}_i)$. Given, $\boldsymbol{\bar{a}} = [\bar{a_i}]_{i=1}^N$, they prove that 
    \begin{equation}
        \sum \limits_{i=1}^N Q_i(\tau_i, a_i, \theta_i) - Q_{tot}(\boldsymbol{\tau},\boldsymbol{a}) + V_{tot}(\boldsymbol{\tau},\boldsymbol{a}) = 
        \begin{cases}
        0 & \boldsymbol{a} = \boldsymbol{\bar{a}} \\ 
        \ge 0 & \boldsymbol{a} \neq \boldsymbol{\bar{a}} 
        \end{cases}
    \end{equation}
    Based on this theory, three networks are built: individual $Q_i$, $Q_{tot}$, and the joint regularizer $V_{tot}$ and three loss functions are demonstrated to train the networks. 
    The local network at each agent is just a regular value-based network, with the local observation which provides the Q-value of all possible actions and runs locally at the execution time. Both $Q_{tot}$ and the regularizer networks use hidden features from the individual value-based network to help sample efficiency.
    In the experimental analysis, the comparisons of QTRAN with VDN and QMIX on Multi-domain Gaussian Squeeze \citep{holmesparker2014exploiting} and modified predator-prey \citep{stone2000multiagent} is provided.

	Within the cooperative setting with the absence of joint reward, the reward-shaping idea can be applied too. Specifically, assume at time step $t$ agent $i$ observes its own local reward $r_i^t$. 
	In this setting, \cite{mguni2018inducing} considers a multi-agent problem in which each agent observes the full state, takes its local action based on a stochastic policy. A general reward-shaping algorithm for the multi-agent problem is discussed and proof for obtaining the Nash equilibrium is provided. In particular, a meta-agent (MA) is introduced to modify the agent's reward functions to get the convergence to an efficient Nash equilibrium solution. The MA initially does not know the parametric reward modifier and learns it through the training. 
	Specifically, MA wants to find the optimal variables $w$ to reshape the reward functions of each agent, though it only observes the corresponding reward of chosen $w$. With a given $w$, the MARL algorithm can converge while the agents do not know anything about the MA function. The agents only observe the assigned reward by MA and use it to optimize their own policy. Once all agents execute their actions and receive the reward, the MA receives the feedback and updates the weight $w$. Training the MA with a gradient-based algorithm is quite expensive, so in the numerical experiments, a Bayesian optimization with an expected improvement acquisition function is used. To train the agents, an actor-critic algorithm is used with a two-layer neural network as the value network. The value network shares its parameters with the actor network, and an A2C algorithm \citep{mnih2016asynchronous} is used to train the actor.
	It is proved that under a given condition, a reward modifier function exists such that it maximizes the expectation of the reward modifier function. In other words, a Markov-Nash equilibrium (M-NE) exists in which each agent follows a policy that provides the highest possible value for each agent. Then, convergence to the optimal solution is proved under certain conditions. To demonstrate the performance of their algorithm, a problem with 2000 agents is considered in which the desire location of agents changes through time.
	
	\section{Consensus} \label{sec:consensus}
	The idea of the centralized critic, which is discussed in Section \ref{sec:centralized_critic}, works well when there exists a small number of agents in the communication network. However, with increasing the number of agents, the volume of the information might overwhelm
    the capacity of a single unit. Moreover, in the sensor-network applications, in which the information is observed across a large number of scattered centers, collecting all this local information to a centralized unit, under some limitations such as energy limitation, privacy constraints, geographical limitations, and hardware failures, is often a formidable task. One idea to deal with this problem is to remove the central unit and allow the agents to communicate through a sparse network, and share the information with only a subset of agents, with the goal of reaching a consensus over a variable with these agents (called neighbors). Besides the numerous real-world applications of this setting, this is quite a fair assumption in the applications of MARL \citep{zhang2018fully,jiang2018graph}. By limiting the number of neighbors to communicate, the amount of communication remains linear in the order of the number of neighbors. In this way, each agent uses only its local observations, though uses some shared information from the neighbors to stay tuned with the network. Further, applying the consensus idea, there exist several works which prove the convergence of the proposed algorithms when the linear approximators are utilized. In the following, we review some of the leading and most recent papers in this area. 
	
	\cite{varshavskaya2009efficient} study a problem in which each agent has a local observation, executes its local policy, and receives a local reward. A tabular policy optimization agreement algorithm is proposed which uses Boltzmann's law (similar to soft-max function) to solve this problem. The agreement (consensus) algorithm assumes that an agent can send its local reward, a counter on the observation, and the taken action per observation to its neighbors. The goal of the algorithm is to maximize the weighted average of the local rewards. In this way, they guarantee that each agent learns as much as a central learner could, and therefore converges to a local optimum.
	
	In \cite{kar2013distributed, kar2013cal} the authors propose a decentralized multi-agent version of the tabular Q-learning algorithm called $\mathcal{QD}$-learning. In this paper, the global reward is expressed as the sum of the local rewards, though each agent is only aware of its own local reward. In the problem setup, the authors assume that the agents communicate through a time-invariant un-directed weakly connected network, to share their observations with their direct neighbors. All agents observe the global state and the global action, and the goal is to optimize the network-averaged infinite horizon discounted reward. The $\mathcal{QD}$-learning works as follows. Assume that we have $N$ agents in the network and agent $i$ can communicate with its neighbors $\mathcal{N}_i$. This agent stores $Q_i(s,a)$ values for all possible state-action pairs. Each update for the agent $i$ at time $t$ includes the regular Q-learning plus the deviation of the Q-value from its neighbor as below:
		\begin{align}
		Q_i^{t+1}({s,a}) = Q_i^t({s,a})&+\alpha_{s,a}^t\left(r{_i}(s_t,a_t)+\gamma\min_{a'\in\mathcal{A}}Q{_i}^t({s_{t+1},a'})-Q{_i}^t({s,a})\right)\nonumber\\
		&{-\beta_{s,a}^t\sum_{j\in\mathcal{N}_i^t}\bigg(Q_i^t({s,a})-Q_j^t({s,a})\bigg)},
		\end{align}
		where $\mathcal{A}$ is the set of all possible actions, and $\alpha_{s,a}^t$ and $\beta_{s,a}^t$ are  the stepsizes of the $\mathcal{QD}$-learning algorithm. 
	It is proved that this method converges to the optimal Q-values in an asymptotic behavior under some specific conditions on the step-sizes. In other words, under the given conditions, they prove that their algorithm obtains the result that the agent could achieve if the problem was solved centrally.
	
	 In \cite{DAC_Pennesi_2010} a distributed Actor-Critic (D-AC) algorithm is proposed under the assumption that the states, actions, and rewards are local for each agent; however, each agent’s action does not change the other agents’ transition models. The critic step is performed locally, meaning that each agent evaluates its own policy using the local reward receives from the environment. In particular, the state-action value function is parameterized using a linear function and the parameters are updated in each agent locally using the Temporal Difference algorithm together with the eligibility traces. The Actor step on the other hand is conducted using information exchange among the agents. First, the gradient of the average reward is calculated. Then a gradient step is performed to improve the local copy of the policy parameter along with a consensus step. A convergence analysis is provided, under diminishing step-sizes, showing that the gradient of the average reward function tends to zero for every agent as the number of iterations goes to infinity. In this paper, a sensor network problem with multiple mobile nodes has been considered for testing the proposed algorithm. In particular, there are $M$ target points and $N$ mobile sensor nodes. Whenever one node visits a target point a reward will be collected. The ultimate goal is to train the moving nodes in the grid such that the long-term cumulative discounted reward becomes maximized. They have been considered a $20\times 20$ grid with three target points and sixteen agents. The numerical results prove that the reward improves over time while the policy parameters reach consensus. 
	
	\cite{macua2017diff} propose a new algorithm, called Diffusion-based Distributed Actor-Critic (Diff-DAC) for single and multi-task multi-agent RL.  
	In the setting, there are $N$ agents in the network such that there is one path between any two agents, each is assigned either the same task or a different task than the others,  and the goal is to maximize the weighted sum of the value functions over all tasks.   Each agent runs its own instance of the environment with a specific task, without intervening with the other agents. For example, each agent runs a given cart-pole problem where the pole length and its mass are different for different agents. Basically, one agent does not need any information, like state, action, or reward of the other agents. Agent $i$ learns the policy with parameters $\theta_i$, while it tries to reach consensus with its neighbors using diffusion strategy. In particular, the Diff-DAC trains multiple agents in parallel with different and/or similar tasks to reach a single policy that performs well on average for all tasks, meaning that the single policy might obtain a high reward for some tasks but performs poorly for the others. The problem formulation is based on the average reward, average value-function, and average probability transition. Based on that, they provide a linear programming formulation of the tabular problem and provide its Lagrangian relaxation and the duality condition to have a saddle point. A dual-ascent approach is used to find a saddle point, in which (i) a primal solution is found for a given dual variable by solving an LP problem, and then (ii) a gradient ascend is performed in the direction of the dual variables. These steps are performed iteratively to obtain the optimal solution. 
	Next, the authors propose a practical algorithm utilizing a DNN function approximator. During the training of this algorithm, agent $i$ first performs an update over the weights of the critic as well as the actor network using local information. Then, a weighted average is taken over the weights of both networks which assures these networks reach consensus.  The algorithm is compared with a centralized training for the Cart-Pole game.
	
	In \cite{zhang2018fully} a multi-agent problem is considered with the following setup. There exists a common environment for all agents and the global state $s$ is available to all of them, each agent takes its own local action $a_i$, and the global action $a = [a_1, a_2, \cdots, a_N]$ is available to all $N$ agents, each agent receives $r_i$ reward after taking an action and this reward is visible only in agent $i$. In this setup, the agents can do time-varying random communication with their direct neighbors ($\mathcal{N}$) to share some information. Two AC-based algorithms are proposed to solve this problem. In the first algorithm, each agent has its own local approximation of the Q-function with weights $w_i$, though a fair approximation needs global reward $r_t$ (not local rewards $r_t^i, \forall i=1,\dots,N$). To address this issue, it is assumed that each agent shares parameter $w_i$ with its neighbors, and in this way a consensual estimate of $Q_{w}$ can be achieved. 
	To update the critic, the temporal difference is estimated by 
	\begin{equation*}
	    \delta^t_{i} = r_i^{t+1} - \mu_i^t + Q(s_{t+1}, a_{t+1},w_i^t) -Q(s_t,a_t,,w_i^t), 
	\end{equation*}
	in which 
	\begin{equation}
	\label{eq:moving_average_reward}
	    \mu_i^t = (1-\beta_{wt}^t)\mu_i^t + \beta_{wt} r^{t+1}_i, 
	\end{equation} 
	i.e. the moving average of the agent $i$ rewards with parameter $\beta_{it}$, and the new weights $w_i$ are achieved locally. To achieve the consensus a weighted sum (with weights coming from the consensus matrix) of the parameters of the neighbors' critics are calculated as below:
	\begin{equation}
	    \label{eq:consensus_critic_w_update}
	    w_i^{t+1} = \sum_{j \in \mathcal{N}_i} c_{ij} \tilde{w}_j^t. 
	\end{equation} 
	 This weighted sum provides the new weights of the critic $i$ for the next time step. To update the actor, each agent observes the global state and the local action to update its policy; though, during the training, the advantage function requires actions of all agents, as mentioned earlier.
	In the critic update, the agents do not share any rewards info and neither actor policy; so, in some sense, the agents keep the privacy of their data and policy. However, they share the actions with all agents, so that the setting of the problem is pretty similar to MADDPG algorithm; although, MADDPG assumes the local observation in the actor.  In the second algorithm, besides sharing the critic weights, the critic observes the moving average estimate for rewards of the neighbor agents and uses that to obtain a consensual estimate of the reward. Therefore, this algorithm performs the following update instead of \eqref{eq:moving_average_reward}:
	\begin{equation*}
	    \tilde{\mu}_i^t = (1-\beta_{wt}^t)\mu_i^t + \beta_{wt} r^{t+1}_i, 
	\end{equation*}
	in which $\tilde{\mu}_i^t = \sum_{j \in \mathcal{N}_i} c_{ij} \tilde{\mu}_j^t.$ Note that in the second algorithm agents share more information among their neighbors. From the theoretical perspective, The authors provide a global convergence proof for both algorithms in the case of linear function approximation. In the numerical results, they provide the results on two examples: (i) a problem with 20 agents and |S|=20, (ii) a completely modified version of cooperative navigation \citep{lowe2017multi} with 10 agents and $|S|=40$, such that each agent observes the full state and they added a given target landmark to cover for each agent; so agents try to get closer to the certain landmark.
	They compare the results of two algorithms with the case that there is a single actor-critic model, observing the rewards of all agents, and the centralized controller is updated there. In the first problem, their algorithms converged to the same return value that the centralized algorithms achieve. In the second problem, it used a neural network and with that non-linear approximation and  their algorithms got a small gap compared to the solutions of the centralized version. 

    In \cite{wai2018multi}, a double-averaging scheme is proposed for the task of {\it policy evaluation} for multi-agent problems. The setting is following \cite{zhang2018fully}, i.e., the state is global, the actions are visible to all agents, and the rewards are private and visible only to the local agent. In detail, first, the duality theory has been utilized to reformulate the multi-agent policy evaluation problem, which is supposed to minimize the mean squared projected Bellman error (MSPBE) objective, into a convex-concave with a finite-sum structure optimization problem. Then, in order to efficiently solve the problem, the authors combine the {\it dynamic consensus} \citep{qu2017harnessing} and the SAG algorithm \citep{schmidt2017minimizing}. Under linear function approximation, it is proved that the proposed algorithm converges linearly under some conditions.
	
	\cite{zhang2018networked} consider the multi-agent problem with continuous state and action space. The rest of the setting is similar to the \cite{zhang2018fully} (i.e., global state, global action, and local reward). Again, an AC-based algorithm is proposed for this problem. In general, for the continuous spaces, stochastic policies lead to a high variance in gradient estimation. Therefore, to deal with this issue deterministic policy gradient (DPG) algorithm is proposed in \cite{silver2014deterministic} which requires off-policy exploration. 
    However, in the setting of \cite{zhang2018networked} the off-policy information of each agent is not known to other agents, so the approach used in DPG \citep{silver2014deterministic, lillicrap2015continuous} cannot be applied here. Instead, a gradient update based on the expected policy-gradient (EPG) \citep{ciosek2018expected} is proposed, which uses a global estimate of Q-value, approximated by the consensus update. Thus, each agent shares parameters $w_i$ of each Q-value estimator with its neighbors.  Given these assumptions, the convergence guarantees with a linear approximator are provided and the  performance is compared with a centrally trained algorithm for the same problem.
	
    Following a similar setting as \cite{zhang2018fully}, \cite{suttle2019multi} propose a new distributed off-policy actor-critic algorithm, such that there exists a global state visible to all agents, each agent takes an action which is visible to the whole network, and receives a local reward which is available only locally. The main difference between this work and \cite{zhang2018fully} comes from the fact that the critic step is conducted in an off-policy setting using emphatic temporal differences $ETD(\lambda)$ policy evaluation method \citep{sutton2016emphatic}. In particular, $ETD(\lambda)$ uses state-dependent discount factor ($\gamma$) and state-dependent bootstrapping parameter ($\lambda$). Besides, in this method there exists an interest function $f: \mathcal{S} \to \mathbb{R}^+$ that takes into account the user's interest in specific states. The algorithm steps are as the following: First, each agent performs a consensus step over the critic parameter. Since the behavior policy is different than the target policy for each agent, they apply importance sampling \citep{kroese2012monte} to re-weight the samples from the behavior policy in order to correspond them to the target policy. Then, an inner loop starts to perform another consensus step over the importance sampling ratio. In the next step, a critic update using $ETD(\lambda)$ algorithm is performed locally and the updated weights are broadcast over the network. Finally, each agent performs the actor update using local gradient information for the actor parameter. Following the analysis provided for $ETD(\lambda)$ in \cite{yu2015convergence}, the authors proved the convergence of the proposed method for the distributed actor-critic method when linear function approximation is utilized.

    \cite{zhang2017datadriven} propose a consensus RL algorithm, in which each agent uses its local observations as well as its neighbors within a given directed graph. The multi-agent problem is modeled as a control problem, and a consensus error is introduced. The control policy is supposed to minimize the consensus error while stabilizes the system and gets the finite local cost. A theoretical bound for the consensus error is provided, and the theoretical solution for having the optimal policy is discussed, which indeed needs environment dynamics.
    A practical actor-critic algorithm is proposed to implement the proposed algorithm. The practical version involves a neural network approximator via a linear activation function. The critic measures the local cost of each agent, and the actor network approximates the control policy. The results of their algorithm on leader-tracking communication problem are presented and compared with the known optimal solution. 
	
	In \cite{distributed_macua_2015}, an off-policy distributed policy evaluation algorithm is proposed. In this paper, a linear function has been used to approximate the long-term cumulative discounted reward of a given policy (target policy), which is assumed to be the same for all agents, while different agents follow different policies along the way. In particular, a distributed variant of the Gradient Temporal Difference (GTD) algorithm  \footnote{ GTD algorithm is proposed to stabilize the TD algorithm with linear function approximation in an off-policy setting.}  \citep{GTD2_Sutton_2009} is developed utilizing a primal-dual optimization scenario. In order to deal with the off-policy setting, they have applied the importance sampling technique. The state space, action space, and transition probabilities are the same for every node, but their actions do not influence each other. This assumption makes the problem stationary. Therefore, the agents do not need to know the state and the action of the other agents. Regarding the reward, it is assumed that there exists only one global reward in the problem. First, they showed that the GTD algorithm is a stochastic Arrow-Hurwicz \footnote{Arrow-Hurwicz is a primal-dual optimization algorithm that performs the gradient step on the Lagrangian over the primal and dual variables iteratively} \citep{arrow_hurwicz} algorithm applied to the dual problem of the original optimization problem. Then, inspiring from \cite{chen2012diffusion}, they proposed a diffusion-based distributed GTD algorithm. Under sufficiently small but constant step-sizes, they provide a mean-square-error performance analysis which proves that the proposed algorithms converge to a unique solution. In order to evaluate the performance of the proposed method, a 2-D grid world problem with 15 agents is considered. Two different policies are evaluated using distributed GTD algorithm. It is shown that the diffusion strategy helps the agents to benefit from the other agents' experiences.  

    Considering similar setup as \cite{distributed_macua_2015}, in \cite{multiagent_stankovic_2016} two multi-agent policy evaluation algorithms were proposed over a time-varying communication network. A given policy is evaluated using the samples derived from different policies in different agents (i.e. off-policy). Same as \cite{distributed_macua_2015}, it is assumed that the actions of the agents do not interfere with each other. Weak convergence is provided for both algorithms. 
	
	Another variant of the distributed GTD algorithm was proposed in \cite{PrimalDual_Lee_2018}. Each agent in the network is following a local policy $\pi_i$ and the goal is to evaluate the global long-term reward, which is the sum of the local rewards. In this work, it is assumed that each agent can observe the global joint state. A linear function, that combines the features of the states, is used to estimate the value function. The problem is modeled as a constrained optimization problem (consensus constraint), and then following the same procedure as \cite{distributed_macua_2015}, a primal-dual algorithm was proposed to solve it. A rigorous convergence analysis based on the ordinary differential equation (ODE) \citep{borkar_ode_2000} is provided for the proposed algorithm.  To keep the stability of the algorithm, they add some box constraints over the variables. Finally, under diminishing step-size, they prove that the distributed GTD (DGTD) converges with probability one. One of the numerical examples is a stock market problem, where $N=5$ different agents have different policies for trading the stocks. DGTD is utilized to estimate the average long-term discounted profit of all agents. The results are compared with a single GTD algorithm in the case the sum of the reward is available. The comparison results show that each agent can successfully approximate the global value function.
	
	\cite{cassano2018multi} consider two different scenarios for the policy evaluation task: (i) each agent is following a policy (behavior policy) different than others, and the goal is to evaluate a target policy (i.e. off-policy). In this case, each agent has only knowledge about its own state and reward, which is independent of the other agents' state and reward. (ii) The state is global and visible to all agents, the reward is local for each agent, and the goal is to evaluate the target team policy. They propose a Fast Diffusion for Policy Evaluation (FDPE) for the case with a finite data set, which combines off-policy learning, eligibility traces, and linear function approximation. This algorithm can be applied to  both scenarios mentioned earlier. The main idea here is to apply a variance-reduced algorithm called AVRG \citep{ying2018convergence} over a finite data set to get a linear convergence rate. Further, they modified the cost function to control the bias term. In particular, they use $h$-stage Bellman Equation to derive $H$-truncated $\lambda$-weighted Mean Square Projected Bellman Error ($H\lambda$- MSPBE) compare to the usual cases (e.g. \cite{distributed_macua_2015}) where they use Mean Square Projected Bellman Error (MSPBE). It is shown that the bias term can be controlled through $(H,\lambda)$. Also, they add a regularization term to the cost function, which can be useful in some cases.

    A distributed off-policy actor-critic algorithm is proposed in \cite{zhang_distributed_2019}. In contrast to the \cite{zhang2018fully}, where the actor step is performed locally and the consensus update is proposed for the critic, in \cite{zhang_distributed_2019} the critic is performed locally, and the agents asymptotically achieve consensus on the actor parameter. The state space and action space are continuous and each agent has the local state and action; however, the global state and the global action are visible to all agents. Both policy function and value function are linearly parameterized. A convergence analysis is provided for the proposed algorithm under diminishing step-sizes for both actor and critic steps. The effectiveness of the proposed method was studied on the distributed resource allocation problem.

	\section{Learn to Communicate}\label{sec:learn_to_communicate}
    
    As mentioned earlier in Section \ref{sec:consensus}, some environments allow the communication of agents. The consensus algorithms use the communication bandwidth to pass raw observation, policy weight/gradients, critic weight/gradients, or some combination of them. A different approach to use the communication bandwidth is to learn a communication-action (like a message) to allow agents to be able to send information that they want. In this way, the agent can learn the time for sending a message, the type of the message, and the destination agents. Usually, the communication-actions do not interfere with the environment, i.e., the messages do not affect the next state or reward. 
	\cite{kasai2008learning} proposed one of the first learning to communicate algorithms, in which tabular Q-learning agents learn messages to communicate with other agents in the predator-prey environment. The same approach with a tabular RL is followed in \cite{varshavskaya2009efficient}. 
	Besides these early works, there are several recent papers in this area which utilize the function approximator. In this section, we discuss some of the more relevant papers in this research area. 
	
	In one of the most recent works, \cite{foerster2016learning}\label{foerster2016learning} consider a problem to learn how to communicate in a fully cooperative (recall that in a fully cooperative environment, agents share a global reward) multi-agent setting in which each agent accesses a local observation and has a limited bandwidth to communicate to other agents. Suppose that $\mathcal{M}$ and $\mathcal{U}$ denote message space and action space respectively. In each time-step, each agent takes action $u \in \mathcal{U}$ which affects the environment, and decides for action $m \in \mathcal{M}$ which does not affect the environment and only other agents observe it.  
	The proposed algorithm follows the centralized learning decentralized execution paradigm under which it is assumed in the training time agents do not have any restriction on the communication bandwidth.
	They propose two main approaches to solve this problem. Both approaches use DRQN \citep{hausknecht2015deep} to address partial observability, and disabled experience replay to deal with the non-stationarity. 
	The input of Q-network for agent $i$ at time $t$ includes $o_{i}^{t}$, $h_{i}^{t}$ (the hidden state of RNN), $\{u_{j}^{t-1}\}_j$, and $\{m_{j}^{t-1}\}_j$ for all $j \in \{1,\dots, N\}$.  When parameter sharing is used, $i$ is also added in the input which helps learn specialized networks for agent $i$ within parameter sharing. All of the input values are converted into a vector of the same size either by a look-up table or an embedding (separate embedding of each input element), and the sum of these \emph{same-size vectors} is the final input to the network. The network returns $|\mathcal{M}| + |\mathcal{U}|$ outputs for selecting actions $u$ and $m$. The network includes two layers of GRU, followed by two MLP layers, and the final layer with $|U| + |M|$ representing $|U|$ Q-values. $|M|$ is different on two algorithms and is explained in the following. First they propose \emph{reinforced inter-agent learning} (RIAL) algorithm. To select the communication action, the network includes additional $|M|$ Q-values to select discrete action $m^i_t$.  They also proposed a practical version of RIAL in which the agents share the policy parameters so that RIAL only needs to learn one network. 
	The second algorithm is \emph{differentiable inter-agent learning} (DIAL), in which the message is continuous and the message receiver provides some feedback---in form of gradient---to the message sender, to minimize the DQN loss. In other words, the receiver obtains the gradient of its Q-value w.r.t the received message and sends it back to the sender so that the sender knows how to change the message to optimize the Q-value of the receiver. 
	Intuitively, agents are rewarded for the communication actions, if the receiver agent correctly interprets the message and acts upon that. 
	The network also creates a continuous vector for the communication action so that there is no action selector for the communication action $m^t_i$, and instead, a regularizer unit discretizes it, if necessary. 
	They provide numerical results on the switch riddle prisoner game and three communication games with mnist dataset. The results are compared with the no-communication and parameter sharing version of RIAL and DIAL methods. 

    \cite{jorge2016learning} extend DIAL in three directions: (i) allow communication of arbitrary size, (ii) gradually increase noise on the communication channels to make sure that the agents learn a symbolic language, (iii) agents do not share parameters. They provide the results of their algorithm on a version of "Guess Who?" game, in which two agents, namely "asking" and "answering", participate. The game is around guessing the true image that the answering agent knows, while the asking agent has $n$ images and by asking $n/2$ questions should guess the correct image. The answering agent returns only "yes/no" answers, and after $n/2$ questions the asking agent guesses the target image. The result of their algorithm with different parameters is presented.
    Following a similar line, \cite{lazaridou2016multi} considers a problem with two agents and one round of communication to learn an interpretable language among the sender and receiver agents. The sender receives two images, while it knows the target image, and sends a message to the receiver along with the images. If the receiver guesses the correct image, both win a reward. Thus, they need to learn to communicate through the message. Each individual agent converts the image to a vector using VGG ConvNet \citep{simonyan2014very}. The sender builds a neural network on top of the input vector to select one of the available symbols (values $10$ and $100$ are used in the experiments) as the message. The receiver embeds the message into the same size as of the images' vector, and then through a neural network combines them together to obtain the guess. Both agents use REINFORCE algorithm \citep{williams1992simple} to train their model and do not share their policies with each other. There is not any pre-designed meaning associated with the utilized symbols. Their results demonstrate a high success rate and show that the learned communications are interpretable.
    In another work in this direction, in \cite{das2017learning} a fully cooperative two-agent game is considered for the task of {\it image guessing}. In particular, two bots, namely a questioner bot (Q-BOT) and an answerer bot (ABOT) communicate in natural language and the task for Q-BOT is to guess an unseen image from a set of images. At every round of the game, Q-BOT asks a question, A-BOT provides an answer. Then the Q-BOT updates its information and makes a prediction about the image. The action space is common among both agents consisting of all possible output sequences under a token vocabulary $V$, though the state is local for each agent. For A-BOT the state includes the sequence of questions and answers, the caption provided for the Q-BOT besides the image itself; while the state for Q-BOT does not include the image information. There exists a single reward for both agents in this game. 
    Similar to \cite{simonyan2014very} the REINFORCE algorithm \citep{williams1992simple} is used to train both agents. Note that \cite{jorge2016learning} allow "yes/no" actions within multiple rounds of communication, \cite{lazaridou2016multi} consist of one single round with continuous messages, and \cite{das2017learning} combine them such that multiple rounds of continuous communication are allowed.  
    
    Similarly, \cite{mordatch2018emergence} study a joint reward problem, in which each agent observes locations and communication messages of all agents. Each agent has a given goal vector $g$ accessed only privately (like moving to or gazing at a given location), and the goal may involve interacting with other agents. Each agent chooses one physical action (e.g., moving or gazing to a new location) and chooses one of the $K$ symbols from a given vocabulary list. The symbols are treated as  abstract categorical variables without any predefined meaning and agents learn to use each symbol for a given purpose. All agents have the same action space and share their policies. 
    Unlike \cite{lazaridou2016multi} there is an arbitrary number of agents and they do not have any predefined rules, like speaker and listener, and the goals are not specifically defined such as the correct utterance. 
    The goal of the model is to maximize the reward while creating an interpretable and understandable language for humans. To this end, a soft penalty also is added to encourage small vocabulary sizes, which results in having multiple words to create a meaning. 
    The proposed model uses the state variable of all agents and uses a fully connected neural network to obtain the embedding $\Phi_s$. Similarly, $\Phi_c$ is obtained as an embedding of all messages. Then, it combines the goal of the agent $i$, $\Phi_s$, and $\Phi_c$ through a fully connected neural network
    to obtain $\psi_u$ and $\psi_c$. Then, the physical action $u$ is equal to $\psi_u + \epsilon$ and the communication message is $c \sim G(\psi_c)$, in which $\epsilon \sim N(0,1)$ and $G(c) = -\log \left(-\log(c) \right)$ is a Gumble-softmax estimator \citep{jang2016categorical}. The results of the algorithm are compared with a no-communication approach in the mentioned game. 

	\cite{sukhbaatar2016learning} consider a fully cooperative multi-agent problem in which each agent observes a local state and is able to send a continuous communication message to the other agents. They propose a model, called CommNet, in which a central controller takes the state observations and the communication messages of all agents, and runs  multi-step communications to provide actions of all agents in the output. 
	CommNet assumes that each agent receives the messages of all agents. In the first round, the state observations $s_i$ of agent $i$ are encoded to $h_i^0$, and the communication messages $c_i^0$ are zero. Then in each round $0<t<K$, the controller concatenates all $h_i^{t-1}$ and $c_i^{t-1}$, passes them into function $f(.)$, which is a linear layer followed by a non-linear function and obtains $h_i^t$ and $c_i^t$ for all agents. To obtain the actions, $h_i^K$ is decoded to provide a distribution over the action space. Furthermore, they provide a version of the algorithm that assumes each agent only observes the messages of its neighbor. 
	Note that, compared to \cite{foerster2016learning}, commNet allows multiple rounds of communication between agents, and the number of agents can be different in different episodes. The performance of CommNet is compared with independent learners, the fully connected, and discrete communication over a traffic junction and Combat game from \cite{sukhbaatar2015mazebase}. 
	
	To extend CommNet, \cite{hoshen2017vain} propose Vertex Attention Interaction Network (VAIN), which adds an attention vector to learn the importance weight of each message. Then, instead of concatenating the messages together, the weighted sum of them is obtained and used to take the action. VAIN works well when there are sparse agents who interact with each agent. They compare their solution with CommNet over several environments.

    In \cite{peng2017multiagent} the authors introduce a bi-directional communication network (BiCNet) using a recurrent neural network such that heterogeneous agents can communicate with different sets of parameters. Then a  multi-agent vectorized version of AC algorithm is proposed for a combat game. In particular, there exists two vectorized networks, namely actor and critic networks which are shared among all agents, and each component of the vector represents an agent. The policy network takes the {shared observation} together with the local information and returns the actions for all agents in the network. The Bi-directional recurrent network is designed in a way to be served as a local memory too. Therefore, each individual agent is capable of maintaining its own internal states besides sharing the information with its neighbors. In each iteration of the algorithm, the gradient of both networks is calculated and the weights of the networks get updated accordingly using the Adam algorithm. In order to reduce the variance, they applied the deterministic off-policy AC algorithm \citep{silver2014deterministic}. The proposed algorithm was applied to the multi-agent StarCraft combat game \citep{samvelyan19smac}. It is shown that BiCNet is able to discover several effective ways to collaborate during the game.

	\cite{singh2018learning} consider the multi-agent problem in which each agent has a local reward and observation. An algorithm called Individualized Controlled Continuous Communication Model (IC3Net) is proposed to learn to \emph{what and when to communicate}, which can be applied to cooperative, competitive, and semi-cooperative environments \footnote{ Semi-cooperative environments are those that each agent looks for its own goal while all agents also want to maximize a common goal.}.
    IC3Net allows multiple continues communication cycles and in each round uses a gating mechanism to decide to communicate or not. The local observation $o_i^t$ are encoded and passed to an LSTM model, which its weights are shared among the agents. Then, the final hidden state $h_i^t$ of the LSTM for agent $i$ in time step $t$ is used to get the final policy. 
    A Softmax function $f(.)$ over $h_i^t$ returns a binary action to decide whether to communicate or not. Considering message $c_i^t$ of agent $i$ at time $t$, the action $a_i^t$ and $c_i^{t+1}$ are:
    \begin{align}
        g_i^{t+1} = & f(h^t_i), \\ 
        h_i^{t+1}, l_i^{t+1} = & \texttt{ LSTM}\left(e(o_i^{t}) + c_i^t, h_i^t, l_i^t \right), \\ 
        c_i^{t+1} = & \frac{1}{N-1} C \sum \limits_{i \neq j} h_j^{t+1}  g_j^{t+1}, \\ 
        a_i^t  = & \pi(h_i^t),
    \end{align}
    in which $l_i^t$ is the cell state of the LSTM cell, $C$ is a linear transformator, and $e(.)$ is an embedding function. Policy $\pi$ and gating function $f$ are trained using REINFORCE algorithm \citep{williams1992simple}. In order to analyze the performance of IC3Net, predator-pray, traffic junction \citep{sukhbaatar2016learning}, and StarCraft with explore and combat tasks \citep{samvelyan19smac} are considered. The results are compared with CommNet \citep{sukhbaatar2016learning}, no-communication model, and no-communication model with only global reward.

    In \cite{jaques2018intrinsic}, the authors aim to avoid centralized learning in multi-agent RL problems when each agent observes local state $o_i^t$, takes a local action, and observes local reward $z_i^t$ from the environment. The key idea is to define a reward called intrinsic reward for influencing the other agents' actions. In particular, each agent simulates the potential actions that it can take and measures the effect on the behavior of other agents by having the selected action. Then, the actions which have  a higher effect on the action of other agents will be rewarded more. Following this idea, the reward function $r_i^t=\alpha z_i^t+\beta c_i^t$ is used, where $c_i^t$ is the casual influence reward on the other agents, $\alpha$, and $\beta$ are some trade-off weights. $c_i^t$ is computed by measuring the KL difference in the policy of agent $j$ when $a_i$ is known or is unknown as below: 
    \begin{equation}
    \label{eq:kl_intrinsic}
        c_i^t = \sum_{j \neq i} \left[ D_{KL} \left[ p(a_j^t | a_i^t, o_i^t) || p(a_j^t | o_i^t) \right] \right] 
    \end{equation}
    In order to measure the influence reward, two different scenarios are considered: (i) a centralized training, so each agent observes the probability of another agent’s action for a given  counterfactual, (ii) modeling the other agents' behavior. The first case can be easily handled by the equation \eqref{eq:kl_intrinsic}. In the second case, each agent is learning $p(a_j^t | a_i^t, o_i^t)$ through a separate neural network. In order to train these neural networks, the agents use the history of observed actions and the cross-entropy loss functions. The proposed algorithm is analyzed on harvest and clean-up environments and is compared with an A3C baseline and baseline which allows agents to communicate with each other. This work is partly relevant to \emph{Theory of Mind} which tries to explain the effect of agents on each other in multi-agent settings. To see more details see \cite{rabinowitz2018machine}.

	\cite{das2018tarmac} propose an algorithm, called TarMAC, to learn to communicate in a multi-agent setting, where the agents \emph{learn to what to sent} and also learn to \emph{communicate to which agent}. 
    They show that the learned policy is interpretable, and can be extended to competitive and mixed environments. To make sure that the message gets enough attention from the intended agents, each agent also encodes some information in the continuous message to define the type of the agent that the message is intended for. This way, the receiving agent can measure the relevance of the message to itself. 
    The proposed algorithm follows a centralized training with a decentralized execution paradigm. Each agent accesses local observation, observes the messages of all agents, and the goal is maximizing the team reward $R$, while the discrete actions are executed jointly in the environment. 
    Each agent sends a message including two parts, the signature $(k_i^t \in R^{d_k})$ and the value $(v_i^t \in R^{d_v})$. The signature part provides the information of the intended agent to receive the message, and the value is the message. Each recipient $j$ receives all messages and learns variable $q_j^t \in R^{d_k}$ to receive the messages. Multiplying $q_j^t$ and $k_i^t$ for all $i \in \{1,\dots,N\}$ results in the attention weights of $\alpha_{ij}$ for all messages from agents $i \in \{1,\dots,N\}$. Finally, the aggregated message $c_i^t$ is the weighted sum of the message values, which the weights are the obtained attention values. This aggregated message and the local observation $o_i^t$ are the input of the local actor. Then, a regular actor-critic model is trained which uses a centralized critic. The actor is a single layer GRU layer, and the critic uses the joint actions $\{a_1,\dots,a_N\}$ and the hidden state $\{h_1,\dots,h_N\}$ to obtain the Q-value. Also, the actors share the policy parameters to speed up the training, and multi-round communication is used to increase efficiency. The proposed method (with no attention, no communication version of the algorithm) is compared with SHAPES \citep{andreas2016neural}, traffic junction in which they control the cars, and House3D \citep{wu2018building}  as well as the CommNet \citep{sukhbaatar2016learning} when it was possible.  
    In the same direction of DIAL, \cite{freed2020communication} proposed a centralized training and decentralized execution algorithm based on stochastic message encoding/decoding to provide a discrete communication channel that is mathematically equal to a communication channel with additive noise. The proposed algorithm allows gradients backpropagate through the channel from the receiver of the message to the sender. The base framework of the algorithm is somehow similar to DIAL \citep{foerster2016learning}; although, unlike DIAL, the proposed algorithm is designed to work under (known and unknown) additive communication noises. 

    In the algorithm, the sender agent generates a real-valued message $z$ and passes it to a randomized encoder in which the encoder adds a uniform noise $\epsilon \sim U(-1/M, 1/M)$ to the continues message to get $\tilde{z}$. Then, to get one of the $M=2^C$ possible discrete messages, where it is an integer version of the message by mapping it into $2^C$ possible ranges. The discrete message $m$ is sent to the receiver, where a randomized decoder tries to reconstruct the original continuous message $z$ from $m$. The function uses the mapping of $2^C$ possible ranges to extract message $\hat{z}$, and then deducts a uniform noise to get an approximation of the original message. The uniform noise in the decoder is generated from the same distribution which the sender used to add the noise to the message. It is proved that with this encoder/decoder, $\hat{z} = z + \epsilon'$ that mathematically is equivalent to a system where the sender sends the real-valued message to the receiver through a channel which adds a uniform noise from a known distribution to the message. In addition, they have provided another version of the encoder/decoder functions to handle the case in which the noise distribution is unknown and it is a function of the message and the state variable, i.e., $\hat{m} = P(.|m,\mathcal{S})$.
    
    In the numerical experiments, an actor-critic algorithm is used to train the weights of the networks, in which the critic observes the full state of the system and the actors share their weights. The performance of the algorithm is analyzed on two environments: (i) hidden-goal path-finding, in which in a 2D-grid each agent is assigned with a given goal cell and needs to arrive at that goal, with 5 actions: move into four directions or stay. Each agent observes its location and the goal of other agents. So, they need to find out about the location of other agents and the location of their own goal through communication with other agents, (ii) coordinated multi-agent search, where there are two agents in a 2D-grid problem and they are able to see all goals only when they are adjacent to the goal or on the goal cell. So, the agents need to communicate with others to get some information about their goals. The results of the proposed algorithm are compared with (i) reinforced communication learning (RCL) based algorithm (like RIAL in \cite{foerster2016learning} in which the communication action is treated like another action of the agent and is trained by RL algorithms) with noise, (ii) RCL without noise for all cases, (iii) real-valued message is passed to the agents, (iv) and no-communication for one of the environments. 

	All the papers discussed so far in this section assumed the existence of a communication message and basically, they allow each agent to learn what to send. In a different approach, \cite{jiang2018learning} fix the message type and only allows each agent to decide to start a communication with the agents in its receptive field. 
    They consider the problem that each agent observes the local observation, takes a local action, and receives a local reward.
    The key idea here is that when there is a large number of agents, sharing the information of all agents and communication might not be helpful since it is hard for the agent to differentiate the valuable information from the shared information. In this case, communication might impair learning. To address this issue, an algorithm, called ATOC is proposed in which an attention unit learns when to integrate the shared information from the other agents. 
    In ATOC, each agent encodes the local observation, i.e. $h_i^t = \mu_{I}(o_i^t; \theta_{\mu})$ in which $\theta_{\mu}$ is the weights of a MLP. 
    Every $T$ time step, agent $i$ runs an attention unit with input $h_i^t$ to determine whether to communicate with the agents in its receptive field or not. If it decides to communicate, a communication group with at most $m$ collaborator is created and does not change for $T$ time-steps. Each agent in this group sends the encoded information $h_i^t$ to the communication channel, in which they are combined and $\tilde{h}_i^t$ is returned for each agent $i \in \mathcal{M_i}$, where $\mathcal{M_i}$ is the list of the selected agents for the communication channel. Then, agent $i$ merges $\tilde{h}_i^t$ with $h_i^t$, passes it to the MLP and obtains $a_i^t = \mu_{II}(h_i^t, \tilde{h}_i^t; \theta_{\mu})$.
    Note that one agent can be added in two communication channels, and as a result, the information can be transferred among a larger number of agents.
    The actor and critic models are trained in the same way as the DDPG model, and the gradients of the actor ($\mu_{II}$) are also passed in the communication channel, if relevant. Also, the difference of the Q-value with and without communication is obtained and is used to train the attention unit. Some numerical experiments on the particle environment are done and ATOC is compared with CommNet, BiCNet, and DDPG (ATOC without any communication). Their experiments involve at least 50 agents so that MADDPG algorithm could not be a benchmark.

\section{Other approaches and hybrid algorithms}\label{sec:emerging_topics}

    In this section, we discuss a few recent papers which either combine the approaches in sections \ref{sec:centralized_critic}-\ref{sec:learn_to_communicate} or propose a model that does not quite fit in either of the previous sections. 

\cite{foerster2018multi} consider a problem in which each agent observes a local observation, selects an action which is not known to other agents, and receives a joint reward, known to all agents. Further, it is assumed that all agents access a common knowledge, and all know that any agent knows this information, and each agent knows that all agents know that all agents know it and so on. Also, there might be subgroups of the agents who share more common knowledge, and the  agents inside each group use a centralized policy to take action and each agent plays its own action in a decentralized model.
Typically, subgroups of the agents have more common knowledge and the selected action would result in higher performance than the actions selected by larger groups. So, having groups of smaller size would be interesting. However, there is a computational trade-off between the selecting smaller or larger subgroups since there is numerous possible combination of agents to form smaller groups. 
This paper proposes an algorithm to address this challenge, i.e., divide the agents to a new subgroup or take actions via a larger joint policy.
The proposed algorithm, called MACKRL, provides a hierarchical RL that in each level of hierarchy decides either to choose a joint action for the subgroup or propose a partition of the agents into smaller subgroups.  
This algorithm is very expensive to run since the number of possible jointed-agents increases exponentially and the algorithm becomes intractable.
To address this issue, a pairwise version of the algorithm is proposed in which there are three levels of hierarchies, the first for grouping agents, the second one for either action-selection or sub-grouping, and the last one for action selection.
Also, a Central-V algorithm is presented for training actor and critic networks.  
	
In \cite{shu2018m} a different setting of the multi-agent system is considered. In this problem, a manager along with a set of self-interested agents (workers) with different skills and preferences work on a set of tasks. In this setting, the agents like to work on their preferred tasks (which may not be profitable for the entire project) unless they offered the right bonus for doing a different task.   Furthermore, the manager does not know the skills and preferences (or any distribution of them) of each individual agent in advance.  The problem goal is to train the manager to control the workers by inferring their minds and assigning incentives to them upon the completion of particular goals. The approach includes three main modules. (i) Identification, which uses workers' performance history to recognize the identity of agents. In particular, the performance history of agent $i$ is denoted by $\mathbb{P}_i=\{P_i^t=(\rho^t_{igb}):t=1,2,\cdots,T\}$, where $\rho^t_{igb}$ is the probability of worker $i$ finishes the goal $g$ in $t$ steps with $b$ bonuses. In this module, these matrices are flattened into a vector and encoded to history representation denoted by $h_i$. (ii) Modeling the behavior of agents. A worker's mind is modeled by its performance, intentions, and skills. In mind tracker module, the manager encodes both current and past information to updates its beliefs about the workers. Formally, let's $\Gamma_i^t={(s_i^\tau,a_i^\tau,g_i^\tau,b_i^\tau):\tau=1,2,\cdots,t}$ denotes the trajectory of worker $i$. Then mind tracker module $M$ receives $\Gamma_i^t$ as well as history representation $h_i$ from the first module and outputs $m_i$ as the mind for agent $i$. (iii) Training the manager, which includes assigning the goals and bonuses to the workers. To this end, the manager needs to have all workers as a context defined as $c^{t+1}=C(\{(s_i^{t+1},m_i^t,h_i): i=1,2,\cdots,N\})$, where $C$ pools all workers information. Then utilizing both individual information and the context, the manager module provides the goal policy $\pi^g$ and bonus policy $\pi^b$ for all workers. All three modules are trained using the A2C algorithm. The proposed algorithm is evaluated in two environments: Resource Collection and
Crafting in 2D Minecraft. The results demonstrate that the manager can estimate the workers' mind through monitoring their behavior and motivate them to accomplish the tasks they do not prefer.  

Next, we discuss MARL in a hierarchical setting. To do so, let us briefly introduce the hierarchical RL. In this setting, the problem is decomposed into a hierarchy of tasks such that easy-to-learn tasks are at the lower level of the hierarchy and a strategy to select those tasks can be learned at a higher level of the hierarchy. Thus, in the hierarchical setting, the decisions at the high level are made less frequently than those at the lower level, which usually happens at every step. The high-level policy is mainly focused on long-run planning, which involves several one-step tasks in the low-level of the hierarchy. 
Following this approach, in single-agent hierarchical RL (e.g. \cite{kulkarni2016hierarchical, vezhnevets2017feudal}), a meta-controller at the high-level learns a policy to select the sequence of tasks and a separate policy is trained to perform each task at the low-level.

For the hierarchical multi-agent systems, two possible scenarios are  synchronous and asynchronous. In the synchronous hierarchical multi-agent systems, all high-level agents take action at the same time. In other words, if one agent takes its low-level actions earlier than other agents, it has to wait until all agents finish their low-level actions. This could be a restricted assumption if the number of agents is quite large. On the other hand, there is no restriction on asynchronous hierarchical multi-agent systems. Nonetheless, obtaining high-level cooperation in asynchronous cases is challenging. In the following, we study some recent papers in hierarchical MARL. 

In \cite{tang2018hierarchical} a cooperative problem with sparse and delayed rewards is considered, in which each agent accesses a local observation, takes a local action, { and submit the joint action}  into the environment to get the local rewards. Each agent has some low-level and high-level actions to take such that the problem of the task selection for each agent can be modeled as a hierarchical RL problem.
To solve this problem, three algorithms are proposed: Independent hDQN (Ind-hDQN),  hierarchical Communication networks (hCom), and hierarchical hQmix. Ind-hDQN is based on the hierarchical DQN (hDQN) \citep{kulkarni2016hierarchical} and decomposes the cooperative problem into independent goals and then learns them in a hierarchical manner. In order to analyze Ind-hDQN, first, we describe hDQN---for the single-agent---and then explain Ind-hDQN for multi-agent setting. 
In hDQN, the meta-controller is modeled as a semi-MDP (SMDP) and the aim is to maximize
$$\tilde{r}_t = R(s_{t+\tau} | s_t, g_t) = r_t + \dots + r_{t+\tau},$$ 
where, $g_t$ is the selected goal by the meta-controller and $\tau$ is the stochastic number of periods to achieve the goal. Via $\tilde{r}_t$, a DQN algorithm learns the meta-controller policy. This policy decides which low-level task should be taken at each time step. Then, the low-level policy learns to maximize the goal-dependent reward $\hat{r}_t$. In 
 Ind-hDQN it is assumed that agent $i$ knows local observation $o_i^t$, its meta-controller learns policy $\pi_i(g_i^t | o_i^t)$, and in the low-level it learns policy $\hat{\pi}_i (a_i^t | g_i^t)$ to interact with the environment. The low-level policy is trained by the environment's reward signals $r_i^t$ and the meta-controller's policy is trained by the intrinsic reward $\hat{r}_i^t$. Since Ind-hDQN trains independent agents, it can be applied to both synchronous and asynchronous settings. 

In the second algorithm, named hCom, the idea of CommNet \citep{sukhbaatar2016learning} is combined with Ind-hDQN. In this way, Ind-hDQN's neural network is modified to include the average of the $h^{\text{th}}$ hidden layers of other agents, i.e., it is added as the $(h+1)^{\text{th}}$ layer of each agent. Similar to Ind-hDQN, hCom works for both synchronous and asynchronous settings.  
The third algorithm, hQmix, is based on Qmix \citep{rashid2018qmix} to handle the case that all agents share a joint reward $r_t$. To this end, the Qmix architecture is added to the meta-controller and as a result, the Qmix allows training separated Q-values for each agent. This is possible by learning $Q_{tot}$ as is directed in the Qmix. hQmix only is applicable for synchronous settings, since $Q_{tot}$ is estimated over joint-action of all agents. 
In each of the proposed algorithms, the neural network's weights of the policy are shared among the tasks that have the same input and output dimensions. Moreover, the weights of the neural network are shared among the agents for the low-level policies. Thus, only one low-level network is trained; although, it can be used for different tasks and by all agents. 
In addition, a new experience replay, Augmented Concurrent Experience Replay (ACER), is proposed. ACER saves transition tuple $(o_i^t, g_i^t, \tilde{r}_i^t, \tau, o_i^{t+\tau})$ for meta-controller and saves 
$AE_i(t,\tau) = \{(o_i^{t+k}, g_i^t, \tilde{r}_i^{t+k}, \tau - k, o_i^{t+\tau} )\}_{k=0}^{\tau-1}$ to train the low-level policy. ACER also uses the idea of Concurrent Experience Replay Trajectories (CERTs) \citep{omidshafiei2017deep} such that experiences are stored in the rows of episodes and columns of time steps to ensure availability of concurrent mini-batches. 
They have analyzed their algorithm in Multiagent Trash Collection tasks (an extension of environment \cite{makar2001hierarchical}) and Fever Basketball Defense. 
In the experiments, the low-level learning is done for several homogeneous agents so that it can be considered as a single-agnet learning problem. The results are compared with Ind-DQN and Ind-DDQN with prioritized experience replay \citep{schaul2015prioritized}.

In the literature of MARL, there are several papers which consider the games with Nash equilibrium. In a multi-agent system, Nash equilibrium is achieved, if all agents get their own highest possible value-function and not willing to change it. Here we only discuss a few recent papers since \cite{lanctot2017unified} provide a detailed review of the older papers and \cite{yang2020overview} present a full review from the game theoretical perspective on the multi-agent formulation and MARL algorithms.

\cite{zhang2018scc} discuss the coordination problem in the multi-agent cooperative domains with a fully observable state and continues actions to find the Pareto-optimal Nash-equilibrium. The paper proposes an algorithm named Sample Continuous Coordination with recursive Frequency Maximum Q-Value (SCC-rFMQ) which includes two main parts: (i) given state $s$, a set of discrete actions from the continues set $A_i(s)$ are selected for agent $i$, (ii) the action evaluation and  training the policy are performed. The first phase involves selecting a set of good actions which are seen before while performs exploration. In this way, it follows a Coordination Re-sample (CR) strategy that preserves $n/3$ best previous actions for each agent. The rest of the actions are selected according to the variable probability distribution, i.e. get actions randomly via $N(a_{max}(s), \sigma)$, in which $a_{max}(s)$ is the action that gives maximum Q-value for state $s$. Let use  $a^{\ast}(s)$ to denote the action of the best seen Q-value. If $a_{max}(s) \neq a^{\ast}(s)$, exploration rate $\sigma_i(s)$ is reset to the initial value of 1/3; otherwise if $V(s) < Q_i(s,a_{max})$, CR shrinks $\sigma_i(s)$ with a given rate; and expands it otherwise. Using the new $\sigma_i(s)$, new actions are selected with $N(a_{max}, \sigma)$ and resets $A_i(s)$ and $Q(s,a)$.  
	
   Beside CR, \cite{zhang2018scc} also utilize rFMQ \citep{matignon2012independent}, which extends Q-Learning with the frequency value $F(s,a)$. In this way, in addition to $Q(s,a)$, $Q_{max}(s,a)$ is also obtained and updated through the learning procedure. To select actions, an evaluation-value function $E(s,a)$ is obtained to run greedily such that $E(s,a) = (1-F(s,a)) Q(s,a)+F(s,a)Q_{max}(s,a)$ and $F(s,a)$ estimates the percentage of time that action $a$ results in observing the maximum reward, for state $s$. The estimation of the frequency $F(s,a)$ is recursively updated via a separate learning rate. 
   In order to show the effectiveness of SCC-rFMQ, the results of a climbing game, a two-player game with Nash equilibrium are presented, as well as the boat navigation problem with two actions. SCC-rFMQ is compared with MADDPG, Sequential Monte Carlo Methods (SMC) \citep{lazaric2008reinforcement}, and rFMQ \citep{matignon2012independent}.  
    SMC \citep{lazaric2008reinforcement} itself is an actor-critic algorithm with continues action space. In this algorithm, the actor takes action randomly such that the probability of extraction of each action is equal to the importance weight of the action. Also, the critic approximates an action-value function based on the observed reward of the playing action. Then, based on the function, the actor updates the policy distribution. In this way, for each state, the actor provides a probability distribution over the continuous action space and based on the importance of sampling, the action is selected.

\section{Applications}\label{sec:application}

The multi-agent problem and MARL algorithms have numerous applications in the real world. In this section, we review some of the application-oriented papers, in which the problem is modeled as a multi-agent problem and an MARL is utilized to solve that. The main focus will be on describing the problem and what kind of approach is used to solve the problem so that there are not much of technical details about the problem and algorithm. Table~\ref{tab:applications} provides a summary of some of the iconic reviewed papers in this section. As it is shown, the IQL approach is the most utilized approach in the application papers. For more details about each paper see the corresponding section. 

\begin{table}[]
\begin{adjustbox}{width=1\textwidth}
\begin{tabular}{llll}\hline
\multicolumn{1}{c}{Category} & \multicolumn{1}{c}{ Problem} & \multicolumn{1}{c}{Goal} & \multicolumn{1}{c}{Algorithm}  \\ \hline
Web service & Task scheduling for web services \cite{wang2016multi} & Minimize time and cost & Tabular Q-learning \\
Traffic control &  Control multi traffic signals \cite{prabuchandran2014multi} & Minimize queue on a neighborhood  & Tabular Q-learning \\
Traffic control & Control multi-intersection traffic signal \cite{presslight19} & Minimize total travel time & IQL \\
Traffic control & Control multi-intersection traffic signal \cite{gong2019decentralized} & Minimize cumulative delay & IQL with DDQN \\
Traffic control & Control multi-intersection traffic signals \cite{chu2019multi} & Minimize queue & IAC \\
Traffic control & Control multi-intersection's traffic signal  \cite{wei2019colight} & Minimize queue  & AC with attention \\
Traffic control & Control single and multi-intersection's traffic signal \cite{zheng2019learning} & Minimize queue & IQL with ApeX-DQN \\
Traffic control         & Ride-sharing management \cite{lin2018efficient} & Improve resource utilization & Q-learning and AC \\
Traffic control         & Air-traffic control \cite{brittain2019autonomous} & Conflict resolution & CTDE A2C                                           \\
Traffic control & Bike re-balancing problem \cite{xiao2018distributed}  & Improve trips frequency \& bike usage & tabular Q-learning \\
Resource allocation     & Online resource allocation \cite{wu2011novel,wu2018decentralised}   & maximize the utility of server & tabular Q-learning \\
Resource allocation     & Packet routing in wireless sensor networks \cite{ye2015multi} & Minimize consumed energy & Q-learning \\
Robot path planning     & Multi agent path finding with static obstacles \cite{sartoretti2019primal} & Find the shortest path & IAC with A3C \\
Robot path planning     & Multi Agent path finding  with dynamic obstacles \cite{wang2020mobile} & Find the shortest path  & Double DQN \\
Production systems      & Production control, job-shop scheduling \cite{dittrich2020cooperative} & Minimize average cycle time & DQN \\
Production systems      & Transportation in semiconductor fabrication \cite{ahn2020cooperative} & Minimize retrieval time & AC \\
Image classification    & Image classification with swarm of robots \cite{mousavi2019multiagent} & Minimizing classification error & REINFORCE \\
Stock market      & Liquidating of a large amount of stock  \cite{bao2019multi} & Maximize the liquidation sell value & DDPG \\
Stock market      & Buy or sell stocks  \cite{lee2007multiagent} & Maximize the profit & Q-learning \\
Maintenance planning  & Maintenance management \cite{andriotis2019managing} & Minimize life-cycle cost  & AC based \\ \hline
\end{tabular}
\end{adjustbox}
\caption{A summary of applications for multi-agent problems with MARL algorithms which are reviewed in this paper.}
\label{tab:applications}
\end{table}

\subsection{Web Service Composition} A multi-agent Q-learning algorithm has been proposed in \cite{wang2016multi} for {\it dynamic web service composition} problem. In this problem, there exists a sequence of tasks that need to be done in order to accomplish the web service composition problem. The key idea in this work is to decompose the main task into independent sub-tasks. Then a tabular-based Q-learning algorithm is applied to find the optimal strategy for the sub-task. In addition, to improve the efficiency of the proposed method they proposed the experience sharing strategy. In this case, there exists a supervisor agent, who is responsible to communicate with all other agents to spread the existing knowledge in a particular agent among all other ones.

 \subsection{Traffic Control} \cite{prabuchandran2014multi} propose a tabular Q-learning algorithm to control {\it multiple traffic signals} on neighbor junctions to maximize the traffic flow. Each agent (the traffic light) accesses the local observations, including the number of lanes and the queue length at each lane, decides about the green light duration, and shares the queue length with its neighbors. The cost of each agent is the average of the queue length at its neighbors so that it tries to minimize the queue length of its own queue as well as all its neighbors. The algorithm is compared with two classical approaches in a simulated environment of two areas with 9 and 12 junctions in India. A similar problem with the RL approach is studied in \cite{abdoos2011traffic}. Also, recently \cite{zhang2019cityflow} provided CityFlow, a new traffic-signal environment to be used for MARL researches. 

CityFlow has been used in many traffic signal control problems. In an intersection, there are some predefined sets of phases---determined based on the structure of the intersection---and the goal of the problem can be translated into deciding about the sequence of these phases to minimize the total travel of all vehicles. However, total travel time is not a direct function of state and actions in an intersection, so that usually auxiliary objective functions like minimizing queue length, waiting time, or delay time are considered.
A variety of traffic statistics such as the number of moving/waiting cars in each lane, the queue length, the number of waiting cars in each lane, etc., can be used as the state $s_t$. The action set is usually defined as the set of all possible phases.  Typically, the reward is defined as a combination of several components such as queue length, the waiting time of the cars, intersection pressure, etc. See \cite{wei2019survey} for a detailed review.

\cite{presslight19} consider the multi-intersection traffic signal control problem and propose an IQL type algorithm to solve it. Each intersection is considered as an RL agent, which observes the current phase, the number of cars on the outgoing road, and the number of cars in each segment of the incoming road. The action is to decide about the next active phase, and the reward of each intersection is the negative of the corresponding pressure. 
There is no parameter sharing among the agents, and each agent trains its own weights. Numerical experiments on several synthetic and real-world traffic cases are conducted to show the performance of the algorithm. 
In a similar paper, \cite{gong2019decentralized} consider the same problem and proposes an IQL based algorithm. To obtain the state, first, each intersection is divided into several chunks to build a matrix in which each chunk includes a binary variable indicating the existence of a vehicle. Then, to get the state of each intersection, the matrix of the considered intersection and its upstream and downstream intersections are obtained and concatenated together to mitigate the complexity of the multi-agent problem. The reward is defined as the difference between the waiting times of all vehicles between two consecutive cycles, and the action is the next phase to run. The goal of the model is to minimize the cumulative delay of all vehicles. 
To solve the problem an IQL approach is proposed in which the agents are trained with the double dueling deep Q network \citep{wang2015dueling} algorithm, where a CNN network along with an FC layer is used to obtain the advantage values.  To explore the performance of the algorithm, a commercial traffic simulator, Aimsun Next, is used as the environment and the real-world data from Florida is used in which eight traffic signals are controlled by the proposed algorithm.

Cooperation among the agents plays a pivotal role in the traffic signal control problem since the action for every individual agent will directly influence the other agents. There have been some efforts to remedy this issue. For example, \cite{prashanth2010reinforcement} consider a central controller that watches and controls all other agents. This strategy suffers from the curse of dimensionality.
Another approach is to assume that agents could share their states among the neighbors \citep{arel2010reinforcement}. For example, \cite{chu2019multi} show that sharing local information could be very helpful though keep the algorithm practical. They propose MA2C, a fully cooperative in which each intersection trains an independent advantage actor-critic where it allows sharing the observations and probability simplex of the policy with the neighbor agents. So, the agent has some information about regional traffic and can try to alleviate that rather than focusing on a self-oriented policy to reduce the traffic in a single intersection. To balance the importance of the local and shared information, a spatial discount factor is considered to scale the effect of the shared observation and rewards. Each agent represents its state by the cumulative delay of the first vehicle on the intersection from time $t-1$ to time $t$, as well as the number of approaching cars within a given distance to the intersection. Each agent chooses its next phase and is reward locally by the weighted sum of the queue length along each incoming lane and wait time of cars in each lane of the intersection. MA2C is evaluated on a large traffic grid with both synthetic and real-world traffic data and the results are compared with IA2C, and IQL.

Even though some algorithms considered sharing some information among the agents, still it is not known how important that information are for each agent. To address this issue \cite{wei2019colight} proposed CoLight, an attention-based RL algorithm. Each intersection learns weights for every other agent, and the weighted sum of the neighbors' state is used by each agent. 
In CoLight, the state includes the current one-hot-coded phase as well as the number of cars in each lane of the roads, the action is choosing the next phase, and the goal is to minimize the average queue length on each intersection. 
All intersections share the parameters so that only one network needs to be trained. Synthetic and real-world data-set is used to show the performance of the algorithm, including traffic network in Jinan with 12 intersections, Hangzhou with 16 intersections, and Manhattan with 196 intersections.

None of the mentioned algorithms can address a city-level problem, i.e., thousands of the intersection.
The main issues are (i) the local reward maximization does not guarantee global reward, and gathering the required data is quite challenging, (ii) the action of each intersection affects the others so that the coordination is required to minimize the total travel time. To address these issues, \cite{chachatoward} proposed MPLight, an RL algorithm for large-scale traffic signal control systems.
The state for each agent consists of the current phase and the 12 possible pressure values of the 12 traffic movements. The intersections with a smaller number of movements are zero-padded. The action is to choose one of eight possible phases, and the local reward is the pressure of the intersection. 
A DQN algorithm is proposed with parameter sharing among the intersections. Both synthetic and real-world data-sets are used, from which Manhattan with 2510 signals is the largest analyzed network. 

In \cite{zheng2019learning} an RL-based algorithm, called FRAP, was proposed for the traffic signal control problem. The key property of FRAP is invariancy to symmetric operations such as rotation and flip. Toward this end, two principles of competition are utilized: (i) that larger traffic movement indicates higher demand,  and (ii) the line with higher traffic (demand) is prioritized to the line with lower traffic. In FRAP, the state is defined as the phase and number of vehicles at each lane, the action is choosing the next phase, and the reward is the queue length. 
The proposed model includes three parts: (i) phase demand modeling, which provides the embedded sum of all green-traffic movements on each phase, (ii) phase pair embedding, in which the movement-conflict and phase-demand matrices are built and embedded to the required sizes, and (iii) phase pair competition, which runs two convolutions neural network and then some fully connected layers to obtain the final Q-values to choose the action.

In a related area, \cite{lin2018efficient} consider ride-sharing management problem and propose a customized Q-learning and an actor-critic algorithm for this problem. The algorithms are evaluated on a built simulation with Didi Chuxinghe, which is the largest ride-sharing company in China. The goal is to match  the demand and supply to improve the utilization of transportation resources. 
In a similar direction, \cite{brittain2019autonomous} study the air-traffic control problem with a MARL algorithm to ensure the safe separation between aircraft. 
Each airplane, as a learning agent, shares the state information with $N$ closest agents. Each state includes distance to the goal, speed, acceleration, distance to the intersection, and the distance to the $N$ closest airplanes. Each agent decides about the change of speed from three possible choices and receives a penalty if it is in a close distance of another airplane. The goal of the model is to identify and address the conflicts between air-crafts in high-density intersections. The A2C algorithm is used, while a centralized training decentralized execution approach is followed. The actor and critic network share the weights. BlueSky air traffic control simulator is used as the environment and the results of a case with 30 airplanes are provided. 
 
In \cite{xiao2018distributed}, a distributed tabular Q-learning algorithm was proposed for {\it bike rebalancing} problem. Using distributed RL, they improve the current solutions in terms of frequency of trips and the number of bikes are being moved on each trip. In the problem setup, the action is the number of bikes to move. 
The agents receive a positive reward if the bike stock is within a particular range, and a negative reward otherwise. Also, there exists a negative reward for every bike moves in each hour. Each agent acts independently; however, there exists a controller called a knowledge repository (KR) that shares the learning information among the agents. More specifically, the KR is designed to facilitate the transfer learning \citep{lazaric2012transfer} among the agents. Using only distributed RL the success ratio ( the number of successfully rebalanced stations over the total stations) is improved about $10\%$ to $35\%$. Furthermore, combining with the transfer learning, the algorithm rebalances the network $62.4\%$ better than the one without transfer learning. 

\subsection{Resource Allocation} \cite{zhang2009multi} consider an online {\it distributed resource allocation} problem, such that each agent is a server and observes the information of its neighbors. In each time step, each agent receives a task and has to decide to allocate it locally, or has to pick a neighbor and send it to that neighbor. Once the task is finished, the agent is rewarded with some utility of the task. Due to the communication bandwidth constraints, the number of tasks that each agent can send to its neighbors is limited and the goal is to cooperatively maximize the utility of the whole cluster.   
An algorithm based on tabular Q-learning is proposed and its results are compared with a centralized controller's result. \cite{wu2011novel} also consider a similar problem and propose another value-based algorithm. Moreover, in \cite{wu2018decentralised} a similar problem is studied in which there are $n$ schedulers who dispatches jobs of $k$ customers in $m$ machines. They also propose a value-based algorithm and use a gossip mechanism to transfer utilities among the neighbor agents.
In a similar domain, \cite{ye2015multi} consider a packet routing problem in the wireless sensor networks, model it as a multi-agent problem. In this problem, the sensor data should be sent to a base station to analyze them, though usually each sensory unit has a limited capacity to store data, and there are strict communication bandwidth limits. This problem has several applications in surveillance, video recording, processing and communicating. When a packet is sent from a given sensory unit, the distance to the destination and the size of the packet determines the required energy to send the packet, and one of the goals of the system is to minimize the sum of consumed energy. They propose an MARL algorithm based on the Q-learning in which each agent selects some cooperating neighbors in a given radius and then can communicate with them. The results are compared with several classical algorithms. Within a similar domain, \cite{dandanov2017dynamic} propose an RL algorithm to deal with the antenna tilt angle in mobile antenna, to get a compromise between the mobile coverage and the capacity of the network. The reward matrix of the problem is built and transition probability among the states are known so that optimal value for each state-action is achieved.  

\subsection{Robot Path Planning}
Multi-agent path finding (MAPF) problem is an NP-hard problem \citep{ma2019searching, lavalle2006planning}. The goal is to find the path for several agents in a system with obstacles for going from given source and destinations for each agent. The algorithms to solve MAPF, in general, can be categorized into three classes: coupled, decoupled, and dynamically-coupled methods. The Coupled approaches treat MAPF as a single high-dimensional agent, which results in the exponential growth of complexity. 
On the other hand, decoupled approaches obtain a separate path for each agent, and adjust the paths with the goal of zero-collisions. Decoupled approaches are able to quickly obtain solutions for large problems \citep{leroy1999multiple,van2011reciprocal}. One of the common approaches to path adjustments is "velocity planning" \citep{cui2012pareto, chen2017decentralized}, which modifies the velocity profile of each agent along with its path to avoid collisions. Similarly, priority planning can be used to allow utilizing the fastest path and speed for the agents with higher priority \citep{ma2016multi,cap2013asynchronous}. 
Even though decoupled approaches can be used for a large number of agents, they are not very effective since they consider a small portion of the joint configuration space and search through low-dimensional spaces \citep{sanchez2002using}. Dynamically coupled approaches are proposed to solve these issues. These approached lie between coupled and decoupled approaches. For example, Conflict-Based Search (CBS) finds the optimal or semi-optimal paths without searching in high-dimensional spaces by building a set of constraints and planning for each agent \citep{barer2014suboptimal, sharon2015conflict}. Also, allowing on-demand growth in the search space during planning \citep{wagner2015subdimensional} is another approach. On the other hand, the location of obstacles may be considered static or dynamic \citep{smierzchalski2005path} which results in another two categorizations for algorithms, off-line and on-line planning, respectively. 

\cite{sartoretti2019primal} consider MAPF problem with static obstacles and proposed PRIMAL, an IQL based approach for decentralized MAPF. PRIMAL uses RL and imitation learning (IL) to learn from an expert centralized MAPF planner. With PRIMAL, the agents use RL to learn efficient single-agent path planning, and IL is used to efficiently learn actions that can affect other agents and the whole team's benefit. This eliminates the need for explicit communication among agents during the execution. They considered a grid-world discrete state, and have defined the local state of each agent as the all information of a 10$\times$10-cell block centered where the agent is located, along with the unit vector directing toward the goal. Actions  are four possible moving (if they are allowed) plus staying still, and the reward is a penalty for each move and a higher penalty for staying still. Also, the collision results in a penalty and reaching the goal has a large positive reward. An algorithm based on A3C is used to train each agent locally, which may results in selfish behavior on each agent, leading to locally optimized actions for each agent. To address this issue three methods are proposed: (i) blocking penalty, (ii) imitation learning, (iii) randomized the size and obstacle density of the world. To use IL, a good heuristic algorithm is used to generate trajectories which are used along with the RL generated trajectories to train the model. It is shown that each of the three methods needs to be in the model and removing either of them results in a big loss in the accuracy of the model. The model is compared with the heuristic approaches which access the full observation of the environment. They also implemented PRIMAL on a small fleet of autonomous ground vehicles in a factory mockup. 

\cite{wang2020mobile} proposed globally guided reinforcement learning (G2RL) that uses a reward structure to generalize to any arbitrary environments. G2RL first calls a global guidance algorithm to get a path for each agent and then during the robot motion, a local double deep Q-Learning (DDQN) \citep{van2016deep} based planner is used to generate actions to avoid the static and dynamic obstacles. The global guidance algorithm observes the whole map and all static obstacles. 
To make the RL agent capable of performing enough exploration, a dense reward function is proposed which encourages the agent to explore freely and tries to not force the agent to strictly follow the global guidance. The global motion planning is calculated once and remains the same during the motion. The dynamic obstacles may move in each time-step and their location become known to the agents when they are within a given distance to the agent. Similar to \cite{sartoretti2019primal}, the agent has five possible actions. The goal is to minimize the number of steps which all agent need to go from the start point to the end point. 

To train the RL agent, on each agent the local observation is passed into a transformer function, and the output is passed into a CNN-LSTM-FC network to get the action. In the local observation, the static obstacle, the dynamic obstacle, and the proposed route by the global guidance are depicted with different colors. The agent uses the Double-DQN algorithm to obtain its local actions. 
Since an IQL-based approach is used to train the agent, it can be used in a system with any number of agents. PRIMAL is compared with several central controller type algorithms to demonstrate its performance in different cases.

\subsection{Production Systems}
In \cite{dittrich2020cooperative}, a cooperative MARL is proposed for production control problem. The key idea in this paper is to use a decentralized control system to reduce the problem complexity and add the capability of real-time decision makings. However, this causes local optimal solutions. To overcome this limitation, cooperative behavior is considered for agents. To perform the idea, a central module that contains a deep Q-learning algorithm is considered. This DQN module, plus a decentralized multi-agent system (MAS) communicates to the manufacturing system (i.e., environment). The MAS module consists of two types of agents, namely, order agents and machine agents. The order agents make scheduling decisions based on the work plan, and the machine agents keep the machines information. These agents are able to collect some local data and the order agents have the capability of making some local decisions following the instructions from the DQN module. For each released order $i$, the subsequent $g$ orders are grouped and denoted by $G_i$. The state is defined as the data related to the orders and the action represents the available machine tools for processing particular orders. The reward contains local reward and global reward. The local reward is defined in a way to encourage order agents for choosing the fastest route, where the global reward takes higher values when the deviation between the actual lead time and the target lead time is smaller. The proposed framework is tested on a job shop problem with three processing steps and the results are compared to a capacity-based solution. 

One of the other studied problems with MARL is the overhead hoist transportation (OHT) problem in semiconductor fabrication (FAB) systems. Three main problems exist in FAB systems: (i) The dispatching problem looks for the assignment of idle OHTs to new loads, like moving a bundle of wafers. The goal of this problem is to minimize average delivery time or to maximize the resource utilization ratio for the material handling. On this problem, several constraints like deadlines or job priorities should be considered. (ii) The problem of determining the optimal path of the OHTs moving from a source machine to a destination machine. The goal of this problem is usually to minimize the total travel time or the total tardiness. (iii) The rebalancing problem, which aims to reallocate idle OHTs over different sections of FAB system. The goal is to minimize the time between the assignment of the load to an OHT and the start of the delivery, which is called retrieval time.  
\cite{ahn2020cooperative} consider the third problem and propose a reinforcement learning algorithm based on the graph neural networks to minimize the average retrieval time of the idle OHTs. 
Each agent is considered as the area that contains two machines to perform a process on the semiconductor. Each agent decides to move or not move an idle OHT from its zone to its neighboring zone. The local state of agent $i$ at time $t$ includes (i) the number of the idle OHTs at time $t$ in zone $i$, (ii) the number of working idle OHTs at time $t$ in zone $i$, (iii) and the number of loads waiting on zone $i$ at time $t$. The observation of agent $i$ at time $t$ includes the $s_t^j$ for all neighbor zones $j$ which share their state with zone $i$. Then, a graph neural network is used to embed the observation of each agent to a vector of a given size. In the graph, each node represents a zone, and the node feature is the state of that zone. The edge of the graph is the number of OHTs moving from one node to another. The policy is a function of the embedded observation, including the state of the neighbor zones. An actor-critic algorithm is proposed, in which the policy parameters are shared among all the agents and the critic model uses the global state of the environment. Several numerical experiments are presented to show the effectiveness of the proposed model compared to some heuristic models. 

\subsection{Image Classification} 
In \cite{mousavi2019multiagent} the authors show that a decentralized multi-agent reinforcement learning can be used for image classification. In their framework, multiple agents receive partial observations of the environment, communicate with the neighbors on a communication graph, and relocate to update their locally available information. Using an extension of the REINFORCE algorithm, an algorithm is proposed to update the prediction and motion planning modules in an end-to-end manner. The result on MNIST data-set is provided in which each agent only observes a few pixels of the image and has used an LSTM network to learn the policy. A similar problem and approach is followed in \cite{mousavi2019layered} on the MNIST data-set.

\subsection{Stock Market}
\cite{bao2019multi} consider the liquidation of a large amount of stock in a limited time $T$. The liquidation process usually is done with the cooperation of several traders/brokers and massively impacts the market. The problem becomes more complex when other entities want to liquidate the same stock in the market. This problem can be modeled as a multi-agent system with (i) competitive objectives: when each agent wants to sell its own stock with the highest price, (ii) cooperative objective: when several agents want to cooperatively sell the stock of one customer at the highest price. The problem is modeled as multi-agent system in which each agent has to select to sell $a \in [0,1]$ percent of stocks in each time step. If the agent selects to sell nothing, it takes the risk of dropped prices and at the end of $T$ time periods, the trader has to sell all remaining stocks even with zero price. An adapted version of the DDPG algorithm is proposed to solve this problem. 

In a related problem, \cite{lee2007multiagent} proposed MQ-Trader, which makes buy and sell suggestions in the stock exchange market. 
MQ-Trader consists of four cooperative Q-learning agents: buy and sell signal agents, which determine a binary action for buy/discard or sell/hold, respectively. These agents want to determine the right time to buy or sell stocks. The other agents, buy and sell order agents, decide about the buy price and sell price, respectively. These agents cooperate to maximize profitability. 
The intuition behind this system is to effectively divide the complex stock trading problem into simpler sub-problems. So, each agent needs to learn specialized knowledge for itself, i.e., buy/sell and price decisions. 
The state for the signal agents is represented by a matrix which is filled with a function of long time price history data. On the other hand, the order agents do not need the long history of the price and use the history of prices in the day to determine the price. The action for the order agents is to choose a best-price ratio over the moving average of the price. And, the reward is given as the obtained profit following the action. The performance of the algorithm is analyzed on KOSPI 200 which includes 200 major stocks in the Korea stock exchange market and its results are compared with some existing benchmarks.

\subsection{Maintenance Management}
As an application in civil engineering, \cite{andriotis2019managing} propose a MARL algorithm for efficient maintenance management of structures, e.g. bridges, hospitals, etc. Each structure has $m>1$ components and the maintenance plan has to consider all of them. The goal is to find the optimal maintenance plan for the whole structure, not the optimal policy for separated components.   
It is assumed that all components observe the global state, and a shared reward is known for all agents. An actor-critic algorithm called Deep Centralized Multi-agent Actor-Critic (DCMAC), proposed to solve this problem. 
DCMAC assumes that given the global state, actions of different components are conditionally independent. This way the authors deal with the non-stationary issue in multi-agent systems.
Therefor, a centralized value-function is trained. Also, a centralized actor network outputs a set of actions $\{|A_1|, \dots, |A_m|\}$, each for one component as well as one set of available actions describing the decisions for the subsystem. This algorithm particularly extends the policy gradient algorithm to the cases with a large number of discrete actions. Since the proposed algorithm is in off-policy setting, an important sampling technique is applied to deal with this issue. Under particular valid assumptions on engineering systems, the proposed algorithm can be extended to the case of Partially Observable MPD (POMDP)s. In the numerical experiments, they cover a broad range of engineering systems including (i) a simple stationary parallel series MDP system, (ii) a non-stationary system with k-out-of-n modes in both MPD and POMDP environments, and (iii) a bridge truss system subject to non-stationary corrosion, simulated through an actual nonlinear structural model, in a POMDP environment. The results prove the effectiveness of the proposed algorithm. 


\section{Environments}\label{sec:environments}
Environments are core elements to train any non batch-MARL algorithm. Basically, the environment provides the scope of the problem and different problems/environments have been the motivation for developing the new RL algorithms. 
Interacting with the real world is usually expensive, time-consuming, or sometimes impossible. Thus, using the simulator of environments is the common practice. Using simulators helps to compare different algorithms and indeed provides a framework to compare different algorithms. 
With these motivations, several-single agent environments are developed. 
Among them, Arcade which provides Atari-2600 \citep{bellemare2013arcade}, MoJoCo (simulates the detailed physics movements of human and some animals body) \citep{todorov2012MuJoCoAP}, OpenAI Gym which gathers these together \citep{brockman2016openai}, PyGame Learning Environment (similar to Arcade) \citep{tasfi2016PLE}, OpenSim (builds musculoskeletal structures of human) \citep{seth2011opensim}, DeepMind Lab (3D navigation and puzzle-solving) \citep{beattie2016deepmind}, ViZDoom (3D Shooting and Navigation Doom using only the visual information) \citep{Kempka2016ViZDoom}, Malmo (based on Minecraft) \citep{johnson2016malmo}, MINOS (Home indoor 3D Navigation) \citep{savva2017minos}, House3D (3D Navigation in indoor area) \citep{wu2018building}, and MazeLab \citep{mazelab} just are few to mention. Either of these environments at least include standard step and reset functions, such that {\tt env.reset()} returns a random initial state and {\tt env.step($a_t$)} returns {\tt $s_{t+1}, r_t, d_t,$ dict} in which {\tt dict} is some additional information. This is a general structure which makes it possible to reproduce a given trajectories with a given policy.

In the multi-agent domain, there is a smaller number of available environments. In addition, there is a broad range of possible settings for sharing information among different agents. For example, some environments involve communication actions, some share the joint reward, some share the global state and each of these cases need special algorithms and not all of the algorithms in Sections \ref{sec:centralized_critic}-\ref{sec:emerging_topics} can be applied to each of these problems. Considering these limitations, there is a smaller number of environments in each setting. Among those, StarCraft II \citep{samvelyan19smac} has achieved a lot of attention in the recent years \citep{foerster2017stabilising, usunier2016episodic, singh2018learning, foerster2018counterfactual,rashid2018qmix,peng2017multiagent}. In this game, each agent only observes its own local information and receives a global reward, though different versions of the game with the globally observable state are also available. Finding dynamic goals also has been a common benchmark for both discrete and continuous action spaces. Multi-agent particle environment \citep{mordatch2017emergence} gathers a list of navigation tasks, e.g., the predator-prey for both discrete and continuous actions. Some of the games allow choosing a communication action too. Harvest-gathering \citep{jaques2018intrinsic} is a similar game with communication-action. Neural MMO \citep{suarez2019neural} provides MMORPGs (Massively Multiplayer Online Role-Playing Games) environment like Pokemon in which agents learn combat and navigation while there is a large population of the same agents with the same goal. 
In the area of traffic management, \cite{zhang2019cityflow} provided CityFlow, a new traffic-signal environment to be used for MARL researches. In addition, \cite{wu2017flow} introduced a framework to control cars within a mixed system of the human-like driver and AI agents. In the same direction as of those real-world like environments, \cite{lussange2021modelling} introduce a simulator of stock market for multi-agent systems. 

In addition to the mentioned environments which are proposed by different papers, few projects gathered some of the environments together to provide a similar framework like OpenAI Gym for the multi-agent system. 
\cite{jiang2019MarlEnv} provide 12 environments navigation/maze-like games. Unity \citep{juliani2018unity} is a platform to develop single/multi-agent games which can be simple grid-world or quite complex strategic games with multiple agents. The resulted game can be used as an environment for training machine learning agents. The framework supports cooperative and competitive multi-agent environments. Unity gives the ability to create any kind of multi-agent environment that is intended; although it is not specifically designed for multi-agent systems. Arena \citep{song2019arena} extends the Unity engine and provides a specific platform to define and build new multi-agent games and scenarios based on the available games. The framework includes 38 multi-agent games from which 27 are new games. New scenarios or games can be built on top of these available games. Designing the games involve a GUI-based configurable social tree, and reward function can be selected from five reward scheme that are proposed to cover most of the possible cases in competitive/cooperative games. Similarly, Ray platform \citep{moritz2018ray} also recently started to support multi-agent environments, and some of the known algorithms are also added in rllib repository \citep{liang2017ray, liang2017rllib}. Ray supports entering/leaving the agents from the problem which is a common case in traffic control suites. 

\section{Potential Research Directions}\label{sec:future_research}
\noindent{\bf Off-policy MARL:} In RL, off-policy refers to learning about a policy (called target policy), while the learning agent is behaving under a different policy (called behavior policy ). This method is of great interest because it can learn about an optimal policy while it explores and also learns about multiple policies only by following a single policy. While the former advantage is helpful in both single and multi-agent settings, the latter one seems to be more fit for the multi-agent setting. In MARL literature, there exist a few algorithms utilizing the off-policy \citep{zhang2018networked, suttle2019multi, distributed_macua_2015}; however, there is still room for research on off-policy in MARL algorithm design, theory, and applications. 

\noindent{\bf Safe MARL:} Safe RL is defined as the training of agent to learn a policy that maximizes the long-term cumulative discounted reward while ensures reasonable performance in order to respect safety constraints and avoid catastrophic situations during the training as well as execution of the policy. The main approaches in safe RL are based on introducing the risk concept in optimality conditions and regulating the exploration process to avoid undesirable actions. There is numerous research in safe RL for single-agent RL. See a comprehensive review in \cite{garcia2015comprehensive}. Nonetheless, the research on safe MARL is very scarce. For example, in \cite{shalev2016safe}, a safe RL algorithm is proposed for Multi-Agent Autonomous Driving. Similarly, in \cite{diddigi2019actor} a framework for the constrained cooperative multi-agent games is proposed, in which to ensure the safety a constraints optimization method is utilized. To solve the problem a Lagrangian relaxation along with the actor-critic algorithm is proposed. 
Given the current limited research on this topic, another straightforward research direction for MARL would be the Safe MARL in order to provide more applicable policies in this setup.

\noindent{\bf Heterogeneous MARL:}
Most of the works we studied above are homogeneous MARL, meaning that all the agents in the network are identical in terms of ability and skill. However, in real-world applications, we likely face a multi-agent problem where agents have different skills and abilities. Therefore, an additional problem here would be how different agents should utilize the other agents' abilities to learn a more efficient policy. As a special case, consider the human-machine interaction. Particularly, humans are able to solve some RL problems very quickly using their experience and cognitive abilities. For example, in a 2D small space, they can find a very good approximation of the shortest path very quickly no matter how complex is the search space. On the other hand, machines have the ability to solve more complex problems in high-dimensional spaces. However, optimality comes at the cost of computational complexity so that oftentimes only a feasible solution is possible.  The question that needs to be answered in this problem is the following: Is it possible to develop MARL algorithms that combine heterogeneous agents' abilities toward maximizing the long-term gain?  Moreover, can this be done in a principled way that comes with performance guarantees? 

\noindent{\bf Optimization in MARL}:
Without a doubt {\it optimization} is an indispensable part of the RL problems. Any progress in optimization methods may lead to more efficient RL and in turn MARL algorithms. In recent years, there has been a flurry of research on designing optimization algorithms for solving complex problems including nonconvex and nonsmooth optimization problems for multi-agent and distributed systems \citep{bianchi2012convergence, di2016next, hong2017prox}. However, the literature of MARL still lacks those algorithms. Future research directions on MARL from the optimization perspective can be divided into two main branches: First, applying the existing optimization algorithms (or adapt them when necessary) to multi-agent problems. For instance, TRPO \citep{schulman2015trust}, which has been shown to be very efficient in single-agent RL problems, might be helpful for multi-agent problems as well. Second, focusing on the theory part of the algorithms. Despite the decent performance of the numerical methods, which utilize the neural networks in MARL, there exists a huge gap between such numerical performance and some kind of convergence analysis. Therefore, this might be the time to think out of the box and focus on the theory part of the neural networks too.  

\adddavood{
\noindent{\bf Inverse MARL:} One of the most vital components of RL is \emph{reward} specification. While in some problems such as games it is trivial, in many other applications pre-specifying reward function is a cumbersome procedure and may lead to poor results. In such circumstances, modeling a skillful agent's behavior is utilized for ascertaining the reward function. This is called \emph{Inverse Reinforcement Learning}. While this area attained significant attention in the single-agent RL problems \citep{arora2021survey}, there is no remarkable contribution regarding the inverse RL for MARL. Therefore, a potential research avenue in MARL would be inverse MARL, how to define relevant components, address the possible challenges,  and extend it to the potential applications.

\noindent{\bf Model-based MARL:} Despite the numerous success stories for model-free RL and MARL, a very typical limitation of these algorithms is sample efficiency. Indeed, these algorithms require a tremendous number of samples to reach good performance. On the other hand, \emph{model-based} RL has been shown to be very successful in a great range of applications \citep{moerland2020model}. For this type of RL algorithms, first, the environment model is learned, and then this model is utilized for prediction and control. In single-agent RL, there exists a significant amount of research regarding the model-based RL methods; see for instance \cite{sun2019model}; however, their extension to MARL has not been explored widely. Therefore, investigating model-based MARL is another worthwhile research direction.
}

\section{Conclusion}\label{sec:conclusion}

    In this review, we categorize MARL algorithms into five groups, namely independent-learners,  fully observable critic,  value function decomposition,  consensus, and learn to communicate. Then we provide an overview of the most recent papers in these classes.  For each paper, first, we have highlighted the problem setup, such as the availability of global state, global action, reward, and the communication pattern among the agents. Then, we presented the key idea and the main steps of the proposed algorithm. Finally, we listed the environments which have been used for evaluating the performance of the algorithm. In addition, among the broad range of applications of MARL for real-world problems, we picked a few representative ones and showed how MARL can be utilized for solving such complicated problems. 

	In Table \ref{table} a summary the of most influential papers in each category is presented. In this summary, we have gathered the general settings of the considered problem and the proposed algorithm to show the gaps and the possible research directions. For example, the table shows that with value decomposition (VD), there is not any research that considers the local states, the local actions, and the local policies. In this table, the third column, {\it com}, shows the communication status, i.e., $0$ means there is no communication among the agents, and $1$ otherwise. In the fourth column, AC means the proposed algorithm is actor-critic based, and $Q$ is used when the proposed algorithm is value-based. In the fifth column, {\it conv} stands for the convergence. Here, $1$ is for the case that the convergence analysis is provided and otherwise it is $0$. In the last three columns tuple (Trn, Exe) stands for (Training and Execution) and G and L are for the {\it global} and {\it local} availability respectively, and determine whether state, action, and policy of each 
    agent is known to other agents. 
    
    {
\renewcommand{\baselinestretch}{1}
\small
\begin{table}[tt]
\begin{tabular}{cccccc|ccc|}
	\cline{7-9}
	&       &    &        &      &      & State        & Action       & Reward \\ \cline{2-9}
	\multicolumn{1}{c}{}     & \multicolumn{1}{|c|}{Reference}  & Com &ComLim & AC/Q & Conv & (Trn, Exe) & (Trn, Exe) & (Trn, Exe) \\ \cline{1-9}
	\multicolumn{1}{|c|}{\multirow{6}{*}{\rotatebox[origin=c]{270}{IQL}}}  & \multicolumn{1}{l|}{ \cite{tan1993multi}}                & 0    &    0     & Q    & 0    & (L,L)        & (L,L)        & (G,G)        \\
	\multicolumn{1}{|c|}{}                     & \multicolumn{1}{l|}{\cite{lauer2000algorithm}}          & 0      &  0     & Q    & 0    & (G,G)        & (L,L)        & (G,G)        \\
	\multicolumn{1}{|c|}{}                     & \multicolumn{1}{l|}{\cite{matignon2007hysteretic}}      & 0      &  0     & Q    & 0    & (G,G)        & (G,L)        & (G,G)        \\
	\multicolumn{1}{|c|}{}                     & \multicolumn{1}{l|}{\cite{tampuu2017multiagent}}        & 0      &  0     & Q    & 0    & (G,G)        & (L,L)        & (G,G)        \\
	\multicolumn{1}{|c|}{}                     & \multicolumn{1}{l|}{\cite{omidshafiei2017deep}}         & 0      &  0     & Q    & 0    & (G,G)        & (G,L)        & (G,G) \\
	\multicolumn{1}{|c|}{}                     & \multicolumn{1}{l|}{\cite{fuji2018deep}}                & 0      &  0     & Q    & 0    & (G,G)        & (G,L)        & (G,G)        \\ \hline
	\multicolumn{1}{|c|}{\multirow{12}{*}{\rotatebox[origin=c]{270}{Fully Observable Critic}}} & \multicolumn{1}{l|}{ \cite{wang2019r}}                   & 1    &  1       & AC   & 0    & (G,L)        & (G,L)        & (L,L)        \\
	\multicolumn{1}{|c|}{}                     & \multicolumn{1}{l|}{ \cite{foerster2018counterfactual}}  & 0    &   0    & AC   & 0    & (G,L)        & (G,L)        & (G,G)        \\
	\multicolumn{1}{|c|}{}                     & \multicolumn{1}{l|}{\cite{ryu2018multi}}                & 0     &   0     & AC   & 0    & (G,L)        & (G,L)        & (L/G,L/G)    \\
	\multicolumn{1}{|c|}{}                     & \multicolumn{1}{l|}{ \cite{sartoretti2019distributed}}   & 0    &   0    & AC   & 0    & (G,G)        & (L,L)        & (L,L)        \\
	\multicolumn{1}{|c|}{}                     & \multicolumn{1}{l|}{\cite{chu2017parameter}}           & 0      &   0    & AC   & 0    & (L,L)        & (L,L)        & (L/G,L/G)    \\
	\multicolumn{1}{|c|}{}                     & \multicolumn{1}{l|}{\cite{yang2018cm3}}                & 0      &   0    & AC   & 0    & (L,L)        & (L,L)        & (L,L)        \\
	\multicolumn{1}{|c|}{}                     & \multicolumn{1}{l|}{\cite{jiang2018graph}}              & 1     &   0    & Q    & 0    & (L,L)        & (L,L)        & (L,L)        \\
	\multicolumn{1}{|c|}{}                     & \multicolumn{1}{l|}{\cite{iqbal2018actor}}              & 1     &   0    & AC   & 0    & (G,L)        & (G,L)        & (G,L)        \\
	\multicolumn{1}{|c|}{}                     & \multicolumn{1}{l|}{\cite{yang2018mean}}                & 1     &   0    & AC/Q & 1    & (G,G)        & (G,L)        & (G,L)        \\
	\multicolumn{1}{|c|}{}                     & \multicolumn{1}{l|}{\cite{kim2019learning}}            & 1      &   1   & AC   & 0    & (G,L)        & (G,L)        & (G,G)        \\
	\multicolumn{1}{|c|}{}                     & \multicolumn{1}{l|}{\cite{lowe2017multi}}              & 0      &   0    & AC   & 0    & (G,L)        & (G,L)        & (L,L)        \\
	\multicolumn{1}{|c|}{}                     & \multicolumn{1}{l|}{ \cite{mao2018modelling}}          & 0      &   0     & AC   & 0    & (G,L)        & (G,L)        & (L,L)        \\
	\hline
	\multicolumn{1}{|c|}{\multirow{3}{*}{\rotatebox[origin=c]{270}{VD}}}  & \multicolumn{1}{l|}{\cite{rashid2018qmix}}              & 0   &    0      & Q    & 0    & (G,L)        & (G,L)        & (G,G)        \\
	\multicolumn{1}{|c|}{}                     & \multicolumn{1}{l|}{\cite{sunehag2018value}}            & 0     &  0     & Q    & 0    & (L,L)        & (L,L)        & (G,G)        \\
	\multicolumn{1}{|c|}{}                     & \multicolumn{1}{l|}{ \cite{mguni2018inducing}}          & 0     &  0     & Q    & 1    & (L,L)        & (L,L)        & (G,G)     \\   
	\multicolumn{1}{|c|}{}                     & \multicolumn{1}{l|}{ \cite{son2019qtran}}                & 0    &  0      & Q    & 1    & (L,L)        & (L,L)        & (G,G)        \\ \hline
	\multicolumn{1}{|c|}{\multirow{10}{*}{\rotatebox[origin=c]{270}{Consensus}}} & \multicolumn{1}{l|}{ \cite{zhang2018fully}}              & 1     &  0      & AC   & 1    & (G,G)        & (G,L)        & (L,L)        \\
	\multicolumn{1}{|c|}{}                     & \multicolumn{1}{l|}{\cite{kar2013cal}}                  & 1    &  0      & Q    & 1    & (G,G)        & (L,L)        & (L,L)        \\
	\multicolumn{1}{|c|}{}                     & \multicolumn{1}{l|}{ \cite{PrimalDual_Lee_2018}}       & 1     &  0      & Q    & 1    & (G,G)        & (L,L)        & (L,L)        \\
	\multicolumn{1}{|c|}{}                     & \multicolumn{1}{l|}{ \cite{distributed_macua_2015}}    & 1     &  0      & Q    & 0    & (L,L)        & (L,L)        & (G,G)        \\
	\multicolumn{1}{|c|}{}                     & \multicolumn{1}{l|}{\cite{macua2017diff}}               & 1    &  0       & AC   & 0    & (L,L)        & (L,L)        & (L,L)        \\
	\multicolumn{1}{|c|}{}                     & \multicolumn{1}{l|}{\cite{cassano2018multi} }           & 1    &  0      & Q    & 1    & (L/G,L/G)    & (L,L)        & (L/G,L/G)    \\
	\multicolumn{1}{|c|}{}                     & \multicolumn{1}{l|}{ \cite{zhang2018networked}}          & 1   &  0       & AC   & 1    & (G,G)        & (G,L)        & (L,L)        \\
	\multicolumn{1}{|c|}{}                     & \multicolumn{1}{l|}{\cite{zhang_distributed_2019}}    & 1      &  0    & AC   & 1    & (G,G)        & (G,G)        & (L,L)        \\ \hline
	\multicolumn{1}{|c|}{\multirow{7}{*}{\rotatebox[origin=c]{270}{Learn to comm}}}  & \multicolumn{1}{l|}{\cite{varshavskaya2009efficient}}   & 1      &  0     & AC   & 1    & (L,L)        & (L,L)        & (L,L)        \\
	\multicolumn{1}{|c|}{}                     & \multicolumn{1}{l|}{\cite{peng2017multiagent} }         & 1    &  0      & AC   & 0    & (G,G)        & (G,G)        & (L,L)        \\
	\multicolumn{1}{|c|}{}                     & \multicolumn{1}{l|}{\cite{foerster2016learning} }       & 1    &  0      & Q    & 0    & (L,L)        & (L,L)        & (G,G)        \\
	\multicolumn{1}{|c|}{}                     & \multicolumn{1}{l|}{\cite{sukhbaatar2016learning} }     & 1    &  0      & AC   & 0    & (L,L)        & (L,L)        & (G,G)        \\
	\multicolumn{1}{|c|}{}                     & \multicolumn{1}{l|}{\cite{singh2018learning}}           & 1    &  0       & AC   & 0    & (L,L)        & (L,L)        & (L,L)        \\
	\multicolumn{1}{|c|}{}                     & \multicolumn{1}{l|}{ \cite{lazaridou2016multi}}          & 1   &  0        & AC   & 0    & (G,G)        & (L,L)        & (G,G)        \\
	\multicolumn{1}{|c|}{}                     & \multicolumn{1}{l|}{ \cite{das2017learning}}             & 1   &  0        & AC   & 0    & (L,L)        & (G,G)        & (L,L)        \\ \cline{1-9} 
\end{tabular}
\vspace{.2cm}
\caption{The proposed algorithms for MARL and the relevant setting. AC stands for all actor-critic and policy gradient-based algorithms, Q represents any value-based algorithm, \adddavood{Com stands for communication, Com = 1 means the agents communicate directly, and Com = 0 means otherwise, ComLim stands for communication bandwidth limit, ComLim = 1 means there is a limit on the bandwidth, and ComLim = 0 means otherwise, Conv stands for convergence, and Conv = 1 means there is a convergence analysis for the proposed method, and Conv = 0 means otherwise, }
the tuple (Trn,Exe) shows the way that state, reward, or action are shared in (training, execution), e.g., (G,L) under state means that during the training the state is observable globally and during the execution, it is only accessible locally to each agent. }
\label{table}
\end{table}
}


\bibliographystyle{plainnat}
\bibliography{main}

\end{document}